\documentclass{article}

\usepackage[preprint]{neurips_2026}

\usepackage[utf8]{inputenc} 
\usepackage[T1]{fontenc}    
\usepackage{graphicx}       
\usepackage{hyperref}       
\usepackage{url}            
\usepackage{booktabs}       
\usepackage{amsfonts}       
\usepackage{nicefrac}       
\usepackage{microtype}      
\usepackage{xcolor}         

\PassOptionsToPackage{numbers,square,sort&compress}{natbib}
\setcitestyle{numbers,square,comma}
\usepackage{wrapfig}
\usepackage{multirow}
\usepackage{bbm}
\usepackage[super]{nth}
\usepackage{amsmath}
\usepackage{algorithm}
\usepackage{listings}
\usepackage{array}

\def\emojielephant{\includegraphics[width=1em]{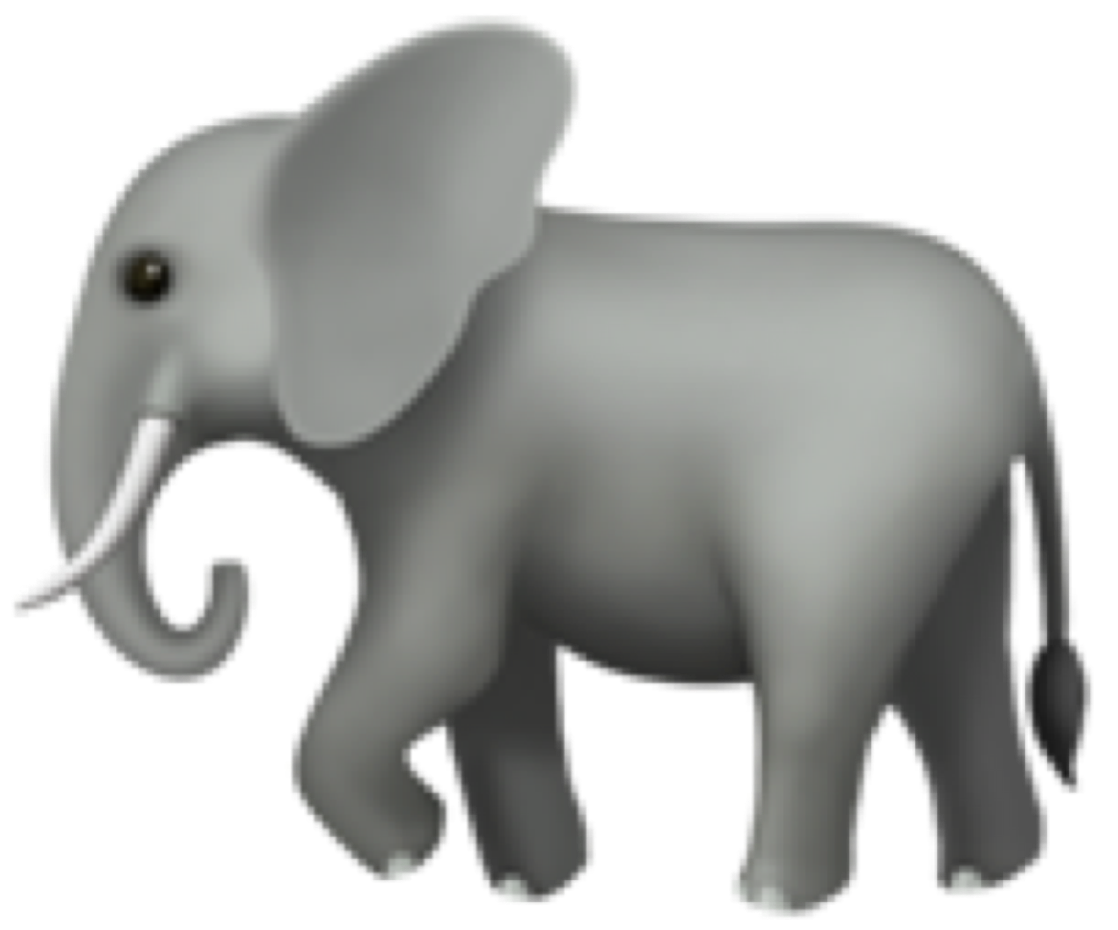}}

\usepackage{linguex}

\definecolor{vecO}{RGB}{230,146,57}   
\definecolor{vecB}{RGB}{0,0,255}
\definecolor{vecG}{RGB}{27,158,119}
\definecolor{vecP}{RGB}{123,50,148}

\title{A Unifying Perspective on Language Model Representations: From Filler-Role Structure to Mechanistic Interpretability}

\author{%
  Zhang Enyan\\
  Department of Computer Science\\
  Yale University\\
  New Haven, CT\\
  \texttt{enyan.zhang@yale.edu}\\
  \And
  R. Thomas McCoy \\
  Department of Linguistics \& Wu Tsai Institute\\
  Yale University\\
  New Haven, CT\\
  \texttt{tom.mccoy@yale.edu}\\
}

\begin{document}

\maketitle

\begin{abstract}
A wide range of methods have been proposed for interpreting language models, delivering important insights into their inner workings. However, different methods and their resulting insights stand in relative isolation: what could the underlying structure of language models be, such that they give rise to all our interpretations? In this work, we propose using Tensor Product Representations (TPRs; \cite{Smolensky1990TPRs}) as a unifying hypothesis. 
TPRs give a concrete proposal for how compositional structure could be represented in vector space --- as filler-role bindings. We show, both mathematically and empirically, that TPRs can unify several prior interpretability methods: additive analogies, linear probing, sparse autoencoders, and activation patching. Mathematically, we show that these methods can all be derived from TPRs. Empirically, we apply the derivations to a range of different models --- from small toy models to LLMs --- to construct instances of each of the above interpretability methods; these constructed variants perform comparably to their standard variants. We view this work as a step toward what interpretability will ideally provide: a unified account of the nature of neural networks, corroborated not just by individual observations but also by an explanation of the connections between them.\footnote{Code is available at \href{https://github.com/Rock-Z/Unifying-Interp-TPRs}{\texttt{https://github.com/Rock-Z/Unifying-Interp-TPRs}}}
\end{abstract}

\section{Introduction}

In a classic story, several people are sent into a completely dark room containing an elephant. None of them have encountered an elephant before. One feels the elephant’s trunk and concludes that elephants are like snakes. Another feels a leg and concludes that elephants are like trees. Another feels a tusk and concludes that elephants are like spears, and the others make observations based on an ear, a side, and a tail (which are like a fan, a wall, and a rope, respectively). Taken together, these observations leave a puzzle: What sort of entity could the elephant be? What underlying structure could explain all of the properties --- the snake-like properties, the tree-like properties, the spear-like properties, etc.? The answer, of course, is a structure like this: \emojielephant

\begin{figure}
    \centering
    \includegraphics[width=.63\linewidth]{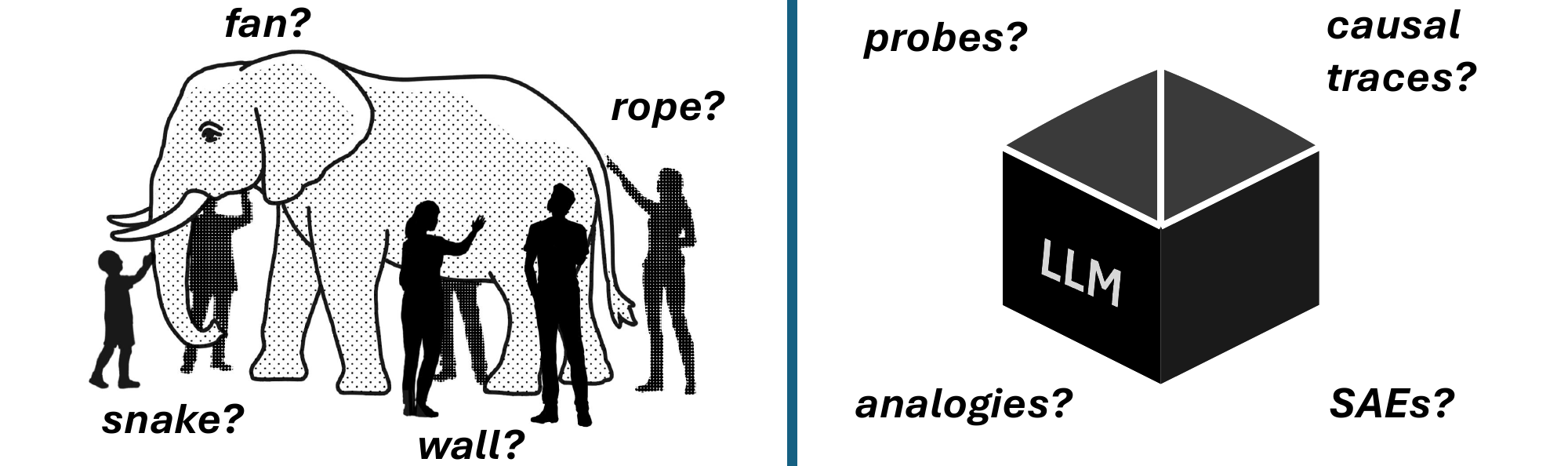}
    \caption{\textbf{Left:} Partial observations of an elephant point to a variety of properties---e.g., the trunk is snake-like and the ears are fan-like. All these partial observations can together be explained by 
    the elephant's overall structure. \textbf{Right:} Similarly, different language model interpretability methods expose consistent but partial descriptions of the same underlying model. Here we ask what the model's underlying structure could be such that it supports all of these observations.}
    \label{fig:elephant}
\end{figure}

A similar situation is found when trying to understand neural network representations: interpretability work has made many important discoveries about neural network representations, but it is not obvious what structure underlies these observations. For instance, these representations support additive analogies \citep{mikolov2013efficientestimationwordrepresentations, pennington-etal-2014-glove}; they can be linearly decoded \cite{ettinger-etal-2016-probing, alain2018understandingintermediatelayersusing, tenney2019bertrediscoversclassicalnlp, hewitt-manning-2019-structural}; they can be decomposed into linear combinations of concepts \cite{cunningham2023sparseautoencodershighlyinterpretable, rajamanoharan2024improvingdictionarylearninggated, gao2024scalingevaluatingsparseautoencoders}; and they can be edited (e.g., via activation patching) in ways that modify behavior \cite{geiger2021causal, wang2022interpretabilitywildcircuitindirect, meng2023locatingeditingfactualassociations, goldowskydill2023localizingmodelbehaviorpath, kramár2024atpefficientscalablemethod}. All of these findings provide windows into the underlying structure of the representations being analyzed. What might the nature of this structure be, such that it can explain all of the properties --- the analogical properties, the decodability properties, etc. --- that have been observed? An answer to this question would move the field toward a more complete picture of neural networks by unifying insights from different methods, behaviors, and types of models (Figure~\ref{fig:elephant}).

We posit that an account of a network’s representations should ideally be both precise and explanatory: 

\setlength{\Extopsep}{4pt}
\setlength{\Exlabelwidth}{0.3em}
\setlength{\SubExleftmargin}{1.35em}
\renewcommand{\firstrefdash}{} 

\ex. \textbf{Precise:} The account should provide an interpretable closed-form equation capturing the structure of the representations, such that representations can be constructed independently of the network being analyzed.\label{desideratum:precise}

\ex. \textbf{Explanatory:} The account should provide explanations of how representations can be analyzed and edited in ways that are compatible with empirical observations. \label{desideratum:explanatory}

In this paper, we propose using Tensor Product Representations (TPRs; \cite{Smolensky1990TPRs}), a formalism for encoding information in vector space, as a unifying account of neural network representations. Building on this formalism, we make the following contributions:
\begin{enumerate}
    \item We show that the closed-form equation provided by TPRs  can be used to approximate the hidden states of a variety of neural networks across different tasks, corroborating prior work \citep{mccoy2019rnnsimplicitlyimplementtensor,Soulos_2020,mccoy2026discover} and extending to new types of networks.
    \item Mathematically, we derive several existing interpretability methods from TPR operations, unifying them under the same structural assumption.
    \item Empirically, we apply the mathematical derivations to construct additive analogies, linear probes, sparse autoencoders, and activation patching from a single TPR approximation, finding performance that is comparable to applying these techniques in the standard ways.
\end{enumerate}
The first contribution shows that the TPR formalism is precise (desideratum \ref{desideratum:precise}), and the second and third contributions show that it is explanatory (desideratum \ref{desideratum:explanatory}).
This work shows that a formal theory of representational structure based on TPRs can connect multiple interpretability results, moving the field toward a more unified understanding of the structure of neural representations.

\section{Background and related work}

This paper brings together two areas of literature concerning the structure of vector representations: mechanistic interpretability and vector symbolic architectures. Here we introduce these lines of literature, which are further discussed in Section~\ref{sec:discussion}.

\subsection{Mechanistic interpretability and the Linear Representation Hypothesis}

Interpretability research aims to reverse engineer the representations and algorithms used by neural networks \cite{bau2017networkdissection, cammarata2020thread:,Olah2022MechInterpVariables, nanda2023progressmeasuresgrokkingmechanistic}, and it uses a wide range of methods. The most prominent existing proposal for unifying multiple methods is the Linear Representation Hypothesis (e.g., \cite{park2024linearrepresentationhypothesisgeometry}), which posits that information is encoded linearly inside neural networks. The Linear Representation Hypothesis is consistent with, and often assumed by, much interpretability work. Though it is often invoked informally or implicitly, this hypothesis can also be formalized. Below we give a possible formalization:

\ex. \textbf{The Linear Representation Hypothesis} \\
Concept level: $\{\mathcal{C}_1, \mathcal{C}_2, \dots, \mathcal{C}_n\}$\\
Vector level: $c_i = emb(\mathcal{C}_i),\ h = \textstyle\sum_{i=1}^n c_i$

That is, at an abstract, conceptual level, a neural network's hidden state $h$ is viewed as encoding a set of concepts $\mathcal{C}_1$ through $\mathcal{C}_n$. To translate abstract concepts to the vector representation $h$, each concept $\mathcal{C}_i$ is encoded with a vector embedding $emb(\mathcal{C}_i)$ and these embeddings are summed to obtain $h$.  

\subsection{Vector symbolic architectures and Tensor Product Representations}
\label{sec:vsa-tprs}

While the Linear Representation Hypothesis has supported many successful analyses, its expressivity is limited in ways that seem at odds with capturing the full structure of neural network representations. For example, one can train a small network to reverse sequences (e.g., $\texttt{2,1,7} \mapsto \texttt{7,1,2}$; Section~\ref{sec:setup-seqs}). A natural way to encode \texttt{2,1,7} under the linear representation hypothesis is as a set of three concepts: 2, 1, and 7. However, this ``bag-of-numbers'' fails to distinguish \texttt{2,1,7} from reorderings such as \texttt{2,7,1}. A possible fix would be to treat each \emph{position-content} pair as an atomic concept, but this produces a large number of concepts and fails to capture how concepts are related. For larger-scale networks, the issue is even more pronounced: e.g., for natural language, creating an atomic concept for each possible combination of a syntactic slot and a word that could fill that slot would require a number of concepts that seems hopelessly large, yet modern LLMs excel at processing language.

It seems, then, that an alternative account that can represent position-content combinations systematically rather than as atomic units is needed. Fortunately, decades of past research in Vector Symbolic Architectures and related formalisms have produced many approaches that share the same goal \cite{Smolensky1990TPRs, pollack1990recursive, plate1995holographic-HRR, gayler1998multiplicative, Rachkovskij2001SBDR, gallant2013mbat, Kleyko_2022a, Kleyko_2022b}. Here, we focus on one specific approach in this tradition: Tensor Product Representations \cite{smolensky1987analysis}. 

A Tensor Product Representation (TPR) encodes structure by binding a filler (content element) and a role (structural position), where each possible filler or role is represented as a vector embedding. E.g., the sequence \texttt{2,1,7} can be thought of as the set of filler-role pairs ``(2, \nth{1}), (1, \nth{2}), (7, \nth{3})''. To encode filler-role pairs in a TPR, we first define the vocabulary of filler and role embedding vectors, $F = \{f_{0}, f_{1}, f_2, \dots\} \in \mathbb{R}^{d_f}$ and $R = \{r_{\text{1st}}, r_{\text{2nd}}, r_{\text{3rd}}, \dots\} \in \mathbb{R}^{d_r}$. Then, each filler-role pair is composed together using the tensor product\footnote{As we use vectors as filler/role embeddings and no recursive bindings, this is equivalent to vector outer products.} as the binding operation, and all filler-role pairs are aggregated with summation. The TPR representing \texttt{2,1,7} is thus $f_{2}\otimes r_{\text{1st}} + f_{1} \otimes r_{\text{2nd}} + f_7 \otimes r_{\text{3rd}}$. Formally, the encoding for a set of filler-role pairs $\{f_i, r_i\}_{i=1}^n$ is
\begin{equation}\label{eq:tpr-e}
    \texttt{encode}(\{f_i, r_i\}_{i=1}^n) = \sum_{i=1}^nf_i \otimes r_i \in \mathbb{R}^{d_f \times d_r}.
\end{equation}
A TPR preserves the information it encodes: given vectors $\{u_1, u_2, ... u_n\}$ such that $r_i \cdot u_j = \delta_{ij}$, where $\delta$ is the Kronecker delta, right-multiplying $u_j$ with a TPR recovers the filler bound to role $j$ in a process known as unbinding:\footnote{Such unbinding vectors are guaranteed to exist as long as the role vectors are linearly independent; see Appendix~\ref{ap:tpr-unbinding}.} %
\begin{equation}\label{eq:tpr-unbind}
    \sum_{i=1}^n f_i \otimes r_i  u_j = \sum_{i=1}^n f_i r_i^\top u_j =\sum_{i=1}^n f_i \delta_{ij} = f_j.
\end{equation}
More generally, TPRs can be constructed for a wide range of content types by using appropriate types of fillers and roles. For example, natural language can be encoded using words as fillers and syntactic positions (e.g., \textit{subject}) as roles. Moreover, TPR binding and unbinding are simple linear or bilinear operations, making them straightforward to implement.

\subsection{The Tensor Product Representation hypothesis}

The structural expressivity of TPRs motivated \cite{mccoy2019rnnsimplicitlyimplementtensor} and \cite{mccoy2026discover} to hypothesize that neural network representations are structured as TPRs. In this paper, we adopt this hypothesis and term it the Tensor Product Representation hypothesis: 

\ex. \textbf{The Tensor Product Representation Hypothesis} \\
Concept level: $\{(\mathcal{R}_1, \mathcal{F}_1), \dots, (\mathcal{R}_n, \mathcal{F}_n)\}$\\
Vector level: \(\begin{aligned}[t]
r_i &= \mathrm{emb}(\mathcal{R}_i),\;f_i = \mathrm{emb}(\mathcal{F}_i),\; h = \textstyle\sum_{i=1}^n f_i \otimes r_i
\end{aligned}\)

In the next sections, we first provide empirical evidence that TPRs can approximate neural network representations. We then show that this fact can be used to derive and explain a range of interpretability methods.

\section{Approximating neural network representations with TPRs}
\label{sec:tprs-tpe}

\begin{wrapfigure}{r}{0.4\linewidth}
    \centering
    \vspace{-1\baselineskip}
    \includegraphics[width=.95\linewidth]{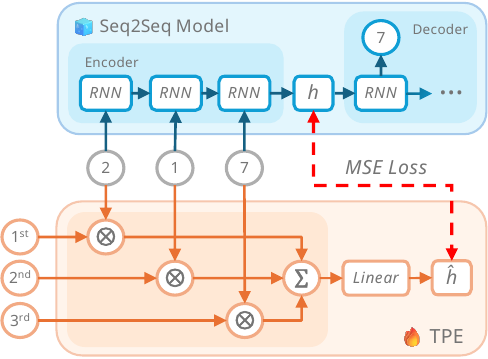}
    \caption{A Tensor Product Encoder training step, where it is trained to approximate representations in a frozen neural network.}
    \label{fig:rnn-tpe}
    \vspace{-1.2\baselineskip}
\end{wrapfigure}

Equation~\ref{eq:tpe} shows the forward pass of a Tensor Product Encoder (TPE; \cite{mccoy2019rnnsimplicitlyimplementtensor}), which operationalizes TPRs into an architecture for approximating neural networks. A TPE takes in a set of filler-role pairs and outputs a linearly transformed TPR:
\begin{equation}\label{eq:tpe}
    \text{TPE}(\{f_i, r_i\}_{i=1}^n) = W\: \text{vec}(\texttt{encode}(\{f_i, r_i\}_{i=1}^n)) + b.
\end{equation}

Figure~\ref{fig:rnn-tpe} presents an example of training a TPE to approximate the encoder hidden state of an RNN: Target activations are obtained from the frozen RNN; then, the TPE receives the input encoded as filler-role pairs;\footnote{We manually specify the filler-role structure of an input in this work, though it can be made unsupervised (\cite{Soulos_2020}).} it is then trained to minimize the mean squared error loss with the target activation.

In interpretability research about representations, there are often two central questions: (i) what concepts are encoded in the model, and (ii) what representational structure/geometry is used to encode those concepts? Our focus is on (ii), such that our experiments would benefit from having little uncertainty about (i)---that is, if we are reasonably confident about what content is present, then we can run experiments that isolate the structural question (ii) because the answer to (i) is already reasonably clear and does not require, e.g., complex unsupervised feature discovery. 
Further, if we know which concepts to expect, we can use that information as the ground truth in probe evaluations, making evaluation easier than what is possible in settings without a clear ground truth.
Accordingly,  we apply the TPE architecture to two domains that each involve a fixed structure: digit sequences with a fixed length of 6 and English sentences with a fixed sentence structure of subject-verb-object (both discussed in the rest of this section). These fixed structures make it straightforward to hypothesize which concepts/features are likely to be encoded. Future work could extend our analyses to domains with more variable structure. 

\subsection{Synthetic sequences}
\label{sec:setup-seqs}

\textbf{Dataset:} We first consider simple algorithmic tasks that involve copying and reversing randomly generated sequences. We generate 50,000 sequences of length 6 with a vocabulary of 20 for each task, split into train/validation/test following an 80:10:10 ratio. Details can be found in Appendix \ref{ap:seqs}.

\textbf{Models:} We train 1-layer encoder-decoder models: recurrent networks (RNNs; \cite{ELMAN1990179}), gated recurrent units (GRUs; \cite{cho-etal-2014-learning}), and long short-term memory networks (LSTMs; \cite{hochreiter1997lstm}) with embedding size 64 and hidden size 256. This setup results in six different conditions, one for each architecture-task combination. All trained networks reach perfect accuracy; Appendix \ref{ap:seqs-model-spec} details training and results.

\textbf{TPE Approximation:} We freeze encoder-decoder models and train Tensor Product Encoders to reconstruct their final encoder hidden state. The TPEs receive, as input, sequences encoded as role-filler pairs of the position (1 through 6) and identity (1 through 20) of each token. The training process and results are detailed in Appendix \ref{ap:seqs-tpe-spec}. Averaged across all six conditions, TPEs achieve an average substitution accuracy of 99.97\% and  $R^2$ of 0.9385; $R^2$ quantifies how well the approximation captures the variation in activation values, and substitution accuracy---the proportion of the time that the decoder produces the correct output when given the approximation rather than the true encoding---quantifies how well the approximation captures the structure that is behaviorally relevant.  The strong approximation metrics are evidence that the encodings being analyzed are indeed structured as linearly-transformed TPRs. 

\subsection{SVO sentences}
\label{sec:setup-svo}

\textbf{Dataset:} Does TPE approximation extend beyond toy tasks to the representation of natural language in state-of-the-art transformers? In order to answer this question, we create a set of structured sentences with the template \texttt{the <subject> will <verb> the <object>.} The subject and object are each sampled from a list of 77 occupations (e.g., ``linguist'', ``chef''), and the verb is sampled from 5 possible verbs. See Appendix \ref{ap:sentences} for more details.

\textbf{Models:} We analyze a set of language models and embedding models across various sizes. Qwen3-8B \citep{yang2025qwen3technicalreport}, OLMo2-13B \citep{olmo20252olmo2furious}, and GPT-OSS-20B \citep{openai2025gptoss120bgptoss20bmodel} (language models), and ModernBERT \citep{warner2024smarterbetterfasterlonger}, EmbeddingGemma \citep{vera2025embeddinggemmapowerfullightweighttext} and Qwen3-Embedding-8B \citep{zhang2025qwen3embeddingadvancingtext} (embedding models). We compute embeddings and representations for these models following their standard usage guidelines. For language models, we analyze the representation of the period token, which \cite{razzhigaev2025llmmicroscopeuncoveringhiddenrole} and \cite{mccoy2026discover} find captures the content of the preceding sentence.

\textbf{TPE Approximation:} We train Tensor Product Encoders to approximate the representations for each model. To construct the input for the TPEs, each sentence is translated into three filler-role pairs for the subject, verb, and object of the sentence. For example, ``the actor will see the chef.'' is translated into the pairs (actor, subject), (see, verb), (chef, object). Appendix~\ref{ap:svo-tpe-spec} details this set of experiments; the TPEs achieve an $R^2$ over 0.90 for all embedding models and an $R^2$ over 0.60 for all LLMs, indicating that they capture most of the variance in these models. The $R^2$ is more modest for the LLMs than for the embedding models, perhaps because the LLM vectors that we analyzed (i.e., the period representations) have not been explicitly optimized to act as sentence representations. 

\section{Additive analogies}

Past work has found additive analogy structures within word embeddings; e.g., \cite{mikolov2013efficientestimationwordrepresentations} observed analogy quartets like ``king $-$ queen $\approx$ man $-$ woman'' where a relationship between two embeddings is represented as a fixed vector difference. Here we first show that representations in our settings also obey such analogies. We then show that additive analogies can be explained, both mathematically and empirically, as an effect of TPR structure.

\subsection{Performing analogy experiments for our setups}
\label{sec:analogies-setup}

\begin{figure}
    \centering
    \includegraphics[width=0.84\linewidth]{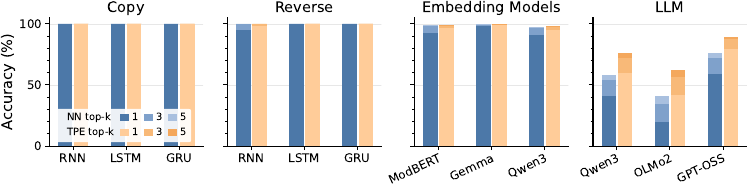}
    \caption{Additive analogy accuracy (\%) evaluated with neural network representations (NN) and TPE-constructed analogy vectors (TPE). Analogies use encoder hidden states for recurrent networks, sentence embeddings for embedding models, and last token hidden state for LLMs. TPE analogies are constructed via Equation~\ref{eq:tpe-analogy}. TPE analogies match or exceed the accuracy of NN analogies.}
    \label{fig:analogies}
\end{figure}

\textbf{Synthetic sequences:} We construct analogy quartets  $A, B, C, D$  for the sequence tasks by substituting tokens from a sequence, such as the following: $\texttt{<2 1 7>} - \texttt{<3 1 0>} + \texttt{<3 5 0>} \approx \texttt{<2 5 7>}$. We sample quartets by substituting up to three positions from an initial sequence (see Appendix~\ref{ap:analogy-seqs}) and report top-$k$ accuracy by ranking the cosine similarity between the analogy vector (left-hand side) and embeddings of all sequences that appeared in an evaluation quartet, and checking if the target (right-hand side) is among the top-$k$.\footnote{An important difference our evaluation and that of  \cite{mikolov2013efficientestimationwordrepresentations} is that we do not discard the embedding of the input sequence, and we use all embeddings as the target comparison set. We make this change to address the concerns raised by \cite{linzen-2016-issues}. See Appendix~\ref{ap:analogy-eval} for the full definition and implementation.}  We find that the embeddings of trained models are  highly structured, with a top-1 accuracy of 99.2\% (Figure~\ref{fig:analogies}).

\textbf{SVO sentences:} We generate quartets $A, B, C, D$ by finding $A$ and $D$ within the test set such that they differ by either subject or object, and then find $B, C$ that shares the same difference (see Appendix~\ref{ap:analogy-svo} for details). We apply the same evaluation procedure to sentence-embedding models and language-model hidden states, comparing the analogy vector against all sentences (Figure~\ref{fig:analogies}).

\subsection{TPRs explain additive analogies}
\label{sec:analogy-tpr}

Additive analogies assume that substituting a component in a representation corresponds to a constant vector offset. Here, we show that this is exactly what the TPR hypothesis predicts. As an example, assume the encoder representation is (approximately) a TPE:
\[
\texttt{<x>} \;=\; W\,\mathrm{vec}\!\:\big(\sum_{s} f_{x,s}\, r_s^\top\big) + b,
\]
where $f_{x,s}$ is the filler embedding at role/position $s$ in input $x$.
Consider two sequences where the difference is changing only the \nth{2} position from \texttt{1} to \texttt{5}:
\[
A=\texttt{<2 1 7>},\qquad D=\texttt{<2 5 7>}.
\]
Substituting in the TPE approximation, we would predict the difference between the two representations to be
$$
\texttt{<2 5 7>} - \texttt{<2 1 7>} = W\,\mathrm{vec}\:\!(f_{5})\, r_{\text{2nd}}^\top - W\,\mathrm{vec}\:\!(f_{1})\, r_{\text{2nd}}^\top = W\,\mathrm{vec}\:\!(f_{5}-f_{1})\, r_{\text{2nd}}^\top,
$$
as all remaining filler-role pairs and the bias terms cancel. For any other pair $(B,C)$ that differs by the same substitution (e.g., $B=\texttt{<3 1 0>}$ and $C=\texttt{<3 5 0>}$), the same cancellation applies. Rearranging terms then yields the standard four-vector analogy ($\texttt{<2 1 7>} - \texttt{<3 1 0>} + \texttt{<3 5 0>} = \texttt{<2 5 7>}$).
Generalizing, if we let $S$ be the set of roles where two TPRs $B$ and $C$ differ, there is
\begin{equation}
\texttt{<}C\texttt{>} - \texttt{<}B\texttt{>}
\;=\;
W\,\mathrm{vec}\:\!\Big(\sum_{s\in S} (f_{C,s}-f_{B,s})\, r_s^\top\Big),
\label{eq:tpe-analogy}
\end{equation}
which predicts that additive analogies hold because they construct filler-role pair differences.

\begin{wrapfigure}{r}{0.4\linewidth}
    \centering
    \includegraphics[width=.95\linewidth]{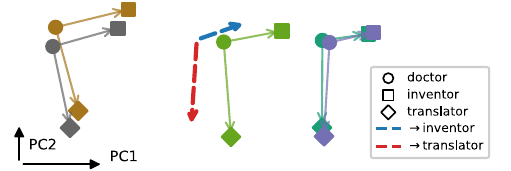}
    \caption{PCA projection of ModernBERT encodings for sentence triplets that differ only in subject. TPE-predicted embedding differences, shown in dashed arrows, closely align with those of sentence triplets.}
    \label{fig:analogy-pca}
\end{wrapfigure}

\subsection{Additive analogies can be constructed from TPEs}

Using the trained TPE for each model, we compute the analogy offset directly with Equation~\ref{eq:tpe-analogy}, without requiring the embeddings for $B$ and $C$, and subtract it from the embedding of $A$. We evaluate with the same ranking-based procedure as Section~\ref{sec:analogies-setup}. Across sequences and SVO sentences (Figure~\ref{fig:analogies}), TPE-constructed analogies match or outperform standard vector analogies despite requiring fewer forward passes (one instead of three) to compute the left-hand side of the analogy.

Figure~\ref{fig:analogy-pca} shows that offsets between ModernBERT sentence embeddings are similar across analogous sentence pairs (cosine similarity \(0.89\pm0.03\)). TPE binding differences match the offsets closely (\(0.94\pm0.01\)), consistent with the account that the offsets represent filler-role binding differences.

\section{Linear probing}
\label{sec:linear-probing}

Linear probes are classifiers (i.e., $\text{logits} = Wh + b$) that predict features from representations. They are widely used to discover implicitly encoded information within neural network representations \cite{ettinger-etal-2016-probing, alain2018understandingintermediatelayersusing, hewitt-manning-2019-structural, tenney2019bertrediscoversclassicalnlp, marks2024geometrytruthemergentlinear}. In this section, we first train linear probes on our task setups. Then, we show that linear probes can be explained by and constructed from TPRs.

\subsection{Performing probing experiments for our setups}

For the synthetic-sequence models, we train one probe per position to take in the encoder’s final hidden state and predict the number at the $i$-th position  (e.g., \texttt{2 1 7 5 10 5} $\mapsto$ \texttt{1} for the ``second'' probe, Appendix~\ref{ap:seqs-probe-spec}). Probes overall perform well, with near-perfect accuracies on RNNs and lower accuracies on a portion of positions of the GRUs/LSTMs. 
For the SVO sentence setting, we train three probes for each model to predict the subject, verb, and object from an input representation (Appendix~\ref{ap:svo-probe-spec}). Trained probes achieve high accuracies across all models and syntactic positions.

\subsection{TPRs explain linear probing}

A probe is a linear map from hidden states to semantic labels, the same as TPR unbinding (Eq.~\ref{eq:tpr-unbind}). Under the TPR hypothesis, probe parameters can therefore be viewed as performing an unbinding mapping, which we can construct explicitly from a trained TPE via three steps: (i) invert the TPE's output layer to recover an approximate TPR encoding from the neural network representation, (ii) unbind from the recovered TPR to extract the filler, and (iii) project the recovered filler embedding onto the filler vocabulary to obtain classification logits. Full derivations are in Appendix~\ref{ap:probe-tpe-derivation}. As all of the steps are linear operations, they can be composed together into the following linear probe weights:
\begin{equation}
    W_{\text{probe}} = F (u_j^\top \otimes I_{d_f}) W^+,\; b_{\text{probe}} = -W_{\text{probe}}\, b.
\end{equation}

\subsection{Linear probes can be constructed from TPEs}

\begin{figure}
    \centering
    \includegraphics[width=.95\linewidth]{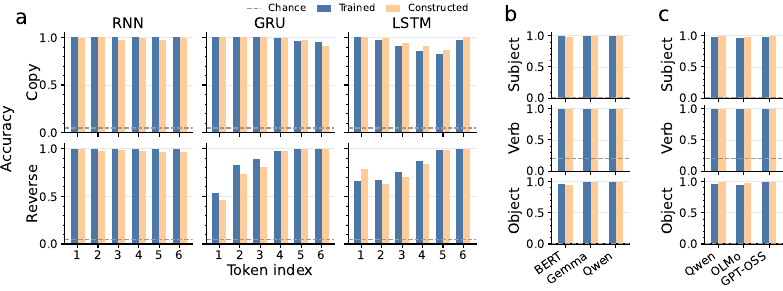}
    \caption{Accuracy comparison between trained linear probes and probes constructed from TPEs. (a) Probes that classify the identity of a token at a specific position (1 through 6) in the sequence; (b) Probes that classify the subject, verb, and object of a sentence given sentence embeddings; (c) Probes that classify subject, verb, and object given a LLM's period token representation at the last layer.}
    \label{fig:probes}
\end{figure}

We utilize the above derivations to construct multiple probes from a single TPE and compare them to trained probes. Constructed probes largely match trained probe accuracy across both digit sequences and SVO sentences (Figure~\ref{fig:probes}), achieving near-perfect accuracies at positions where the trained probes perform well, and similarly worse accuracies at the positions where trained probes perform worse.

Beyond performance comparisons, the constructed probes also lend insight into the relationship between different probes (e.g., subject and object probes). Specifically, the success of the probe construction process (which reuses filler embeddings across positional probes) suggests that different linear probes might be drawing upon shared structural components in the inputs that they decode from --- shared filler representations --- despite differing in which role is unbound.

\section{Sparse autoencoders}
\label{sec:sae}

A sparse autoencoder (SAE) learns sparsely activating features that linearly reconstruct input activations. SAEs often yield semantically meaningful LLM features \cite{cunningham2023sparseautoencodershighlyinterpretable, bricken2023monosemanticity, gao2024scalingevaluatingsparseautoencoders, rajamanoharan2024improvingdictionarylearninggated, templeton2024scaling}, though their efficacy remains debated (e.g., \cite{paulo2025sparseautoencoderstraineddata, tian2025measuringsparseautoencoderfeature, makelov2025towards}). In this section, we show that SAEs can be mathematically and empirically explained by, and derived from, TPRs.

\subsection{Performing SAE experiments for our setups}

We train SAEs on the sentence representations: sentence embeddings for embedding models and period-token final hidden states for language models.
To avoid having to heuristically assign interpretations to features while still enabling tests of how well model representations can be decomposed into SAE features, we follow \cite{makelov2025towards} and \cite{ karvonen2024measuringprogressdictionarylearning} in defining a set of ground-truth concepts against which SAE features can be compared. The ground-truth concepts that we use are pairings of a syntactic position and a word (e.g., subject-actor, verb-see).

We train two types of SAEs: top-$k$ SAEs, and supervised SAEs with an auxiliary loss aligning features to ground-truth concepts (Appendix~\ref{ap:sae-training}). We report reconstruction scores ($R^2$) and \emph{feature quality}, which measures for each concept
the probability the feature activates more on examples where the concept is present than on ones where it is absent. Both SAE types perform well in reconstruction (Table~\ref{tab:sae-comparison}), and supervised SAEs also achieve high feature quality. Top-$k$ SAEs achieve lower feature quality, but this is expected since they are not supervised in what the features should be.

\subsection{SAEs can be derived from TPRs}

A canonical SAE derives feature values from activations with a linear layer followed by a nonlinear activation function, and then reconstructs the input with these features. The first layer, which maps from activations to concept space, can be derived from the TPR unbinding operation in a similar way as was done for linear probes. The weights and biases for an SAE can be constructed from a TPE as follows, where $W_{enc, i}$ and $b_{enc, i}$ give the SAE feature for filler $f_\star$ bound to role $r_j$ (full derivations are in Appendix~\ref{ap:sae-tpe-derivation}): 
\begin{equation}
    W_{enc, i}=W^{+\top}u_j\otimes f_\star,\quad b_{enc, i}=-W_{enc, i}^\top b, \label{eq:sae-feat-onerole}
\end{equation}

\subsection{SAEs can be constructed from TPEs}

\begin{wraptable}[20]{r}{.5\linewidth}
\vspace{-3.5\baselineskip}
\centering
\caption{SAE performance.}
\vspace{7pt}
\small
\setlength{\tabcolsep}{4.5pt}
\begin{tabular}{llrr}
\toprule
Model & SAE type & $R^2$ & Quality (\%) \\
\midrule
ModBERT & Top-$k$ & 0.9803 & 72.52 \\
        & Supervised  & 0.9676 & 99.99 \\
        & TPE-constr. & 0.9289 & 99.77 \\
\midrule
Gemma   & Top-$k$ & 0.9889 & 92.63 \\
        & Supervised  & 0.9866 & 97.26 \\
        & TPE-constr. & 0.9679 & 99.98 \\
\midrule
Qwen3-Emb & Top-$k$ & 0.9785 & 80.84 \\
          & Supervised  & 0.9648 & 99.40 \\
          & TPE-constr. & 0.9163 & 99.53 \\
\midrule
Qwen3-8B & Top-$k$ & 0.9710 & 94.11 \\
         & Supervised  & 0.9979 & 99.78 \\
         & TPE-constr. & 0.9932 & 94.92 \\
\midrule
OLMo2    & Top-$k$ & 0.9843 & 93.97 \\
         & Supervised  & 0.9894 & 99.85 \\
         & TPE-constr. & 0.9822 & 99.32 \\
\midrule
GPT-OSS  & Top-$k$ & 0.9897 & 97.94 \\
         & Supervised  & 0.9909 & 99.45 \\
         & TPE-constr. & 0.9846 & 99.92 \\
\bottomrule
\end{tabular}
\label{tab:sae-comparison}
\end{wraptable}

We construct SAEs by instantiating one feature per filler--role pair, applying Equation~\ref{eq:sae-feat-onerole} and concatenating the resulting features into $W_{enc}, b_{enc}$, and set $W_{dec}$ as the pseudoinverse of $W_{enc}$. Table~\ref{tab:sae-comparison} shows the results of TPE-constructed SAEs compared to trained ones for both embedding-model and language-model representations. Constructed SAEs also reconstruct well (average $R^2$ = $0.96$), though their $R^2$ is lower than that of their trained counterparts; a potential explanation 
is that TPEs are constrained to represent a filler-role pair as either present or absent, while SAEs are more expressive in that they give each feature a continuous weight. Overall, the comparison between trained and constructed SAEs shows that the fact that model encodings can be decomposed via SAEs can be explained via the TPR hypothesis.

\section{Activation patching}

The previous sections make connections between the TPR hypothesis and a range of interpretability methods used for analyzing representations. A natural question to ask, then, is whether model behavior is causally affected by the TPR-like structure that is at the heart of our analyses. In this section, we hypothesize that the edits made by causal intervention methods such as activation patching \cite{wang2022interpretabilitywildcircuitindirect} implicitly work by modifying TPR filler-role binding structure. We show that TPEs can be used to construct activation patching interventions with outcomes comparable to standard activation patching, providing evidence that the TPR structure we have identified in the previous sections is causally implicated in model behavior.  

\subsection{Performing patching experiments for our setups}

We start by considering an activation patching task based on our SVO-structured sentences; the expected behavior is that the network will fill in the blank with the subject of the first sentence (e.g., \textit{linguist} in the \textit{source} example below): 

\emph{Source:} the linguist will see the student. the student will be seen by the \_\_\_\_\_\_ \\ \emph{Destination:} the actor will see the student. the student will be seen by the \_\_\_\_\_\_

\begin{figure}
    \centering
    \includegraphics[width=.5\linewidth]{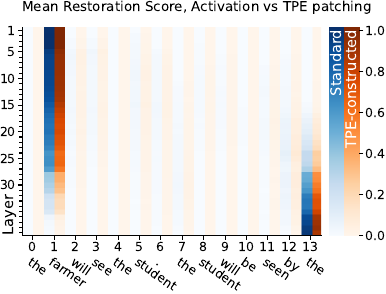}
    \caption{Comparison of restoration scores between standard activation patching (blue hues) and TPE-constructed patching (orange hues), on OLMo-2-13B, across layers and 14 token positions in the prompt. The two methods produce nearly identical results.}
    \label{fig:patching-comparison}
\end{figure}

An activation patching experiment intervenes on the forward pass of the \emph{destination} prompt by replacing the activation at a specific layer and token with the analogous activation from the \emph{source} prompt. If the activation is responsible for the model's output, then the model's response would also change to the response it would produce given the source prompt. We run an exploratory experiment on an autoregressive language model, OLMo-2-13B \cite{olmo20252olmo2furious}. We run token-level activation patching on the same set of three autoregressive language models as seen in Sections \ref{sec:linear-probing} and \ref{sec:sae}, using active/passive SVO prompts. For each trial, we cache all activations from the source prompt, then patch a single layer and token position in the destination prompt. To quantify the effect of a patching trial, we compute \emph{restoration score}, which measures how much the model's behavior is restored via patching.\footnote{Here, we follow best practices by \citet{heimersheim2024useinterpretactivationpatching} and use NNSight \cite{fiottokaufman2024nnsightndifdemocratizingaccess} to run patching experiments.} Consistent with prior work \cite{wang2022interpretabilitywildcircuitindirect}, we find that the effective patch position shifts from the subject to the final token from early to late layers.

\subsection{TPRs explain activation patching}

A patching experiment can only be effective if the activation being patched is systematically structured in a way such that replacing an activation does not collapse the model's performance but instead modifies its behavior accordingly. If the patched representations are TPRs that systematically encode the content of the sentence, then patching would indeed be successful. Further, the source/destination difference would be in the filler bound to the subject role, such that the patched activation should relate to the original activation via a binding difference:
\begin{align}
h_{\textit{dest.}} &= h_{\textit{src.}} - W\!\left(\big(f_{\textit{src,subj}} - f_{\textit{dest,subj}}\big)\otimes r_{\text{subj}}\right)
\label{eq:tpe-activation-patching}
\end{align}
This would predict, in turn, that activation patching does not require two separate forward passes to compute the source and target activations, as $h_\textit{dest}$ can instead be constructed by subtracting the binding difference from $h_{\textit{src.}}$, seen in Equation~\ref{eq:tpe-activation-patching}.

\subsection{Activation patching can be constructed via TPEs}

We train TPEs, one for each layer of the chosen language models, to approximate its activations with the same filler-role structure as the SVO sentences task. All TPEs are trained with filler dimension 128 and role dimension 4, 100 epochs, learning rate 0.002 with cosine scheduling, and batch size 256.

Then, we re-run activation patching on all token positions across all layers, constructing $\hat h_{\textit{dest.}}$ with TPEs following Equation~\ref{eq:tpe-activation-patching}. Crucially, note that the TPE predicts an \emph{activation difference} between the source and destination prompts, and computes the destination prompt representation using that difference instead of another language model forward pass. Figure \ref{fig:patching-comparison} compares the mean restoration score between the two sets of patching experiments across all layers of OLMo and all token positions.

To quantify the similarity of standard activation patching and TPE-constructed patching, we additionally compute the Mean Absolute Error (MAE) and the Pearson Correlation Coefficient ($r$) between the two sets of patching restoration scores. The results are reported below:

\begin{itemize}
    \item OLMo-2-13B: $r = 0.9998$, MAE $= 0.00098$
    \item Qwen3-8B: $r = 0.99999$, MAE $=0.00052$
    \item GPT-OSS-20B: $r = 0.99991$, MAE $= 0.00135$
\end{itemize}

As both metrics and Figure~\ref{fig:patching-comparison} demonstrates, TPE-constructed patching achieves near-identical results as standard activation patching, correctly reproducing both the magnitude of the effect and the shift of information position, suggesting that TPR structure can serve as an explanation for the mechanisms underlying patching.

\subsection{Relationship to other intervention-based methods}

In the above experiment, we explicitly make comparisons between TPE-constructed patching and a two-example patching setup. More broadly, our comparison can also be connected to the assumptions of intervention-based methods that hypothesize that linearly adding and subtracting to representations can causally affect model behaviors in meaningful ways: for example, steering vectors \cite{todd2024functionvectorslargelanguage, hendel-etal-2023-context, pres2024reliableevaluationbehaviorsteering} and causal abstraction \cite{geiger2021causal, pmlr-v236-geiger24a}. This assumption is consistent with the TPR hypothesis, and we connect this assumption to activation patching.

\section{Discussion}
\label{sec:discussion}

We have demonstrated that a range of mechanistic interpretability methods can be explained with a single underlying structure, Tensor Product Representations. In doing so, we also deliver new insights on how these methods operate: for additive analogies, we give a structural hypothesis for the vector differences used in analogies. For linear probing, we show that different probes can be seen as the unbinding of different roles to recover the same set of fillers. For sparse autoencoders, we show that a dictionary of features can contain further internal structure. For activation patching, we show that the difference between patched activations can be explained as a filler-role binding difference. 

\paragraph{Towards structural insights in interpretability:}

A typical interpretability project usually aims to provide an explanation for some behavior of interest within a model. While the explanation can be seen as the main result, it also tests the underlying structural hypotheses. For much of the work that involves interpreting transformers, the hypothesis usually involves some information --- e.g. syntax \cite{hewitt-manning-2019-structural, htut2019attentionheadsberttrack}, semantic objects \cite{wang2022interpretabilitywildcircuitindirect, meng2023locatingeditingfactualassociations}, or truth values \cite{marks2024geometrytruthemergentlinear, hong2025impliesbcircuitanalysis} --- being  represented in some specific way --- e.g., linearly \cite{alain2018understandingintermediatelayersusing, hupkes2018visualisationdiagnosticclassifiersreveal, burns2024discoveringlatentknowledgelanguage, turner2024steeringlanguagemodelsactivation}, in activations \cite{geva2021transformerfeedforwardlayerskeyvalue, cunningham2023sparseautoencodershighlyinterpretable}, or by an attention head \cite{clark2019doesbertlookat, wang2022interpretabilitywildcircuitindirect, singh2024needsrightinductionhead}. 

These structural assumptions are often treated as technical details or simplifications, necessary for the resource and architectural constraints we operate under and justified by prior empirical results. In contrast, we argue that the success of experiments based on these structural assumptions is an important piece of evidence in its own right. In the short term, emphasizing structural insights helps connect results across models and tasks. If successful explanations of models share similar structural properties, then all can be seen as corroborating the same structural hypothesis. Starting with a structural hypothesis then allows for deriving different interpretability methods as downstream results supported by the same structure; e.g., the same TPE approximation can be used to derive multiple analyses. 

As a longer term goal, it also seems that interpretability should provide broad, structural descriptions that persist across individual behaviors and architectures. Here we take inspiration from early biology and cognitive science. Biology grew from its early descriptive stages of individual species and phenomena to finding unifying regularities across domains \cite{mazzarello1999unifyingcelltheory, darwin1964origin}. Cognitive science emerged from the unification of analyses from different disciplines \cite{newell2007computer, miller2003cognitive}. In a similar spirit, we envision that a future ``science of language models'' should also seek to deliver structural insights that answer questions concerning the nature of neural networks with respect to a broad class of models.

\paragraph{Relationship to the Linear Representation Hypothesis:}

The Linear Representation Hypothesis can also be taken as a unifying theory for mechanistic interpretations \citep{park2024linearrepresentationhypothesisgeometry}. TPRs encapsulate this hypothesis but also add greater structure. Thus, TPRs can serve as a more expressive structural hypothesis that can address two notable limitations of the Linear Representation Hypothesis. 

First, filler and role bindings extend beyond the ``bag of concepts'' view to explain the structure used to combine concepts, offering a theory of how position-content information can be compositionally stored in a single activation that the Linear Representation Hypothesis cannot explain. This type of expressivity is not only powerful but also considered essential for a range of cognitively significant capabilities \cite{newell1980physical, fodor1988connectionism}, many of which are now important targets of explanation in language model interpretability. As a concrete demonstration of such expressivity, we perform an exploratory experiment training TPEs on datasets where a portion of the filler-role pairs are held out from the training set, and we show that the approximation generalizes to these held-out examples (Appendix~\ref{ap:holdout}).

The Linear Representation Hypothesis is also at odds with a growing body of empirical work showing that neural network concepts are not always encoded in a one-dimensionally linear manner \citep{csordás2024recurrentneuralnetworkslearn, engels2025languagemodelfeaturesonedimensionally}. TPRs are multi-dimensionally linear 
and can capture structures such as the one \citep{csordás2024recurrentneuralnetworkslearn} describes: indeed, our GRUs and LSTMs in Section~\ref{sec:linear-probing} represent sequences in ways linear probes cannot fully recover, yet the TPEs trained to approximate them can achieve perfect substitution accuracy. 

\section{Conclusion}

We have shown that the Tensor Product Representation formalism can unify four popular interpretability methods. Expanding such unification to additional methods and/or tasks can potentially deliver additional insights into how interpretability findings relate to each other and to the literature on vector symbolic architectures. Another avenue of work would be towards explaining how TPRs can account for complex, varying structures that go beyond the structural expressivity of the Linear Representation Hypothesis (e.g., syntax trees). More broadly, building theories of the shared structure that underlies the full spectrum of analysis techniques is an important goal for interpretability research---a goal that the current paper makes progress toward. 

\section*{Acknowledgments}

We would like to thank Yumeng Yan for the creating the illustration in Figure \ref{fig:elephant}. For helpful discussion, we are grateful to Najoung Kim, Sophie Hao, Arman Cohan, Robert Frank, Alex Lew, the Computational Linguistics at Yale lab, and the Yale Mechanistic Interpretability group. Any errors are our own.

\bibliographystyle{plainnat}
\bibliography{neurips_2026}

@article{Smolensky1990TPRs,
  title = {Tensor Product Variable Binding and the Representation of Symbolic Structures in Connectionist Systems},
  author = {Smolensky, Paul},
  year = {1990},
  journal = {Artificial Intelligence},
  volume = {46},
  number = {1},
  pages = {159--216},
  issn = {0004-3702},
  doi = {10.1016/0004-3702(90)90007-M},
  url = {https://www.sciencedirect.com/science/article/pii/000437029090007M}
}

@inproceedings{mccoy2019rnnsimplicitlyimplementtensor,
title={{RNN}s implicitly implement tensor-product representations},
author={R. Thomas McCoy and Tal Linzen and Ewan Dunbar and Paul Smolensky},
booktitle={International Conference on Learning Representations},
year={2019},
url={https://openreview.net/forum?id=BJx0sjC5FX},
}

@inproceedings{Soulos_2020,
   title={Discovering the Compositional Structure of Vector Representations with Role Learning Networks},
   url={http://dx.doi.org/10.18653/v1/2020.blackboxnlp-1.23},
   DOI={10.18653/v1/2020.blackboxnlp-1.23},
   booktitle={Proceedings of the Third BlackboxNLP Workshop on Analyzing and Interpreting Neural Networks for NLP},
   publisher={Association for Computational Linguistics},
   author={Soulos, Paul and McCoy, R. Thomas and Linzen, Tal and Smolensky, Paul},
   year={2020},
   pages={238–254} }

@misc{alain2018understandingintermediatelayersusing,
title={Understanding intermediate layers using linear classifier probes},
author={Guillaume Alain and Yoshua Bengio},
year={2017},
url={https://openreview.net/forum?id=ryF7rTqgl}
}

@inproceedings{tenney2019bertrediscoversclassicalnlp,
    title = "{BERT} Rediscovers the Classical {NLP} Pipeline",
    author = "Tenney, Ian  and
      Das, Dipanjan  and
      Pavlick, Ellie",
    editor = "Korhonen, Anna  and
      Traum, David  and
      M{\`a}rquez, Llu{\'i}s",
    booktitle = "Proceedings of the 57th Annual Meeting of the Association for Computational Linguistics",
    month = jul,
    year = "2019",
    address = "Florence, Italy",
    publisher = "Association for Computational Linguistics",
    url = "https://aclanthology.org/P19-1452/",
    doi = "10.18653/v1/P19-1452",
    pages = "4593--4601"
}

@inproceedings{cunningham2023sparseautoencodershighlyinterpretable,
title={Sparse Autoencoders Find Highly Interpretable Features in Language Models},
author={Robert Huben and Hoagy Cunningham and Logan Riggs Smith and Aidan Ewart and Lee Sharkey},
booktitle={The Twelfth International Conference on Learning Representations},
year={2024},
url={https://openreview.net/forum?id=F76bwRSLeK}
}

@inproceedings{
wang2022interpretabilitywildcircuitindirect,
title={Interpretability in the Wild: a Circuit for Indirect Object Identification in {GPT}-2 Small},
author={Kevin Ro Wang and Alexandre Variengien and Arthur Conmy and Buck Shlegeris and Jacob Steinhardt},
booktitle={The Eleventh International Conference on Learning Representations },
year={2023},
url={https://openreview.net/forum?id=NpsVSN6o4ul}
}

@misc{heimersheim2024useinterpretactivationpatching,
      title={How to use and interpret activation patching}, 
      author={Stefan Heimersheim and Neel Nanda},
      year={2024},
      eprint={2404.15255},
      archivePrefix={arXiv},
      primaryClass={cs.LG},
      url={https://arxiv.org/abs/2404.15255}, 
}

@inproceedings{
meng2023locatingeditingfactualassociations,
title={Locating and Editing Factual Associations in {GPT}},
author={Kevin Meng and David Bau and Alex J Andonian and Yonatan Belinkov},
booktitle={Advances in Neural Information Processing Systems},
editor={Alice H. Oh and Alekh Agarwal and Danielle Belgrave and Kyunghyun Cho},
year={2022},
url={https://openreview.net/forum?id=-h6WAS6eE4}
}

@inproceedings{
park2024linearrepresentationhypothesisgeometry,
title={The Linear Representation Hypothesis and the Geometry of Large Language Models},
author={Kiho Park and Yo Joong Choe and Victor Veitch},
booktitle={Causal Representation Learning Workshop at NeurIPS 2023},
year={2023},
url={https://openreview.net/forum?id=T0PoOJg8cK}
}

@inproceedings{smolensky1987analysis,
  title={Analysis of distributed representation of constituent structure in connectionist systems},
  author={Smolensky, Paul},
  booktitle={Neural Information Processing Systems},
  year={1987}
}

@misc{mikolov2013efficientestimationwordrepresentations,
      title={Efficient Estimation of Word Representations in Vector Space}, 
      author={Tomas Mikolov and Kai Chen and Greg Corrado and Jeffrey Dean},
      year={2013},
      eprint={1301.3781},
      archivePrefix={arXiv},
      primaryClass={cs.CL},
      url={https://arxiv.org/abs/1301.3781}, 
}

@inproceedings{pennington-etal-2014-glove,
    title = "{G}lo{V}e: Global Vectors for Word Representation",
    author = "Pennington, Jeffrey  and
      Socher, Richard  and
      Manning, Christopher",
    editor = "Moschitti, Alessandro  and
      Pang, Bo  and
      Daelemans, Walter",
    booktitle = "Proceedings of the 2014 Conference on Empirical Methods in Natural Language Processing ({EMNLP})",
    month = oct,
    year = "2014",
    address = "Doha, Qatar",
    publisher = "Association for Computational Linguistics",
    url = "https://aclanthology.org/D14-1162/",
    doi = "10.3115/v1/D14-1162",
    pages = "1532--1543"
}

@article{plate1995holographic-HRR,
  title={Holographic reduced representations},
  author={Plate, Tony A},
  journal={IEEE Transactions on Neural networks},
  volume={6},
  number={3},
  pages={623--641},
  year={1995},
  publisher={IEEE}
}

@ARTICLE{Rachkovskij2001SBDR,
  author={Rachkovskij, D.A.},
  journal={IEEE Transactions on Knowledge and Data Engineering}, 
  title={Representation and processing of structures with binary sparse distributed codes}, 
  year={2001},
  volume={13},
  number={2},
  pages={261-276},
  doi={10.1109/69.917565}}

@article{gayler1998multiplicative,
  title={Multiplicative binding, representation operators \& analogy (workshop poster)},
  author={Gayler, Ross W},
  year={1998}
}

@article{gallant2013mbat,
  title={Representing objects, relations, and sequences},
  author={Gallant, Stephen I and Okaywe, T Wendy},
  journal={Neural computation},
  volume={25},
  number={8},
  pages={2038--2078},
  year={2013},
  publisher={MIT Press}
}

@article{pollack1990recursive,
  title={Recursive distributed representations},
  author={Pollack, Jordan B},
  journal={Artificial Intelligence},
  volume={46},
  number={1-2},
  pages={77--105},
  year={1990},
  publisher={Elsevier}
}

@article{mazzarello1999unifyingcelltheory,
  title={A unifying concept: the history of cell theory},
  author={Mazzarello, Paolo},
  journal={Nature cell biology},
  volume={1},
  number={1},
  pages={E13--E15},
  year={1999},
  publisher={Nature Publishing Group}
}

@book{darwin1964origin,
  title={On the origin of species: A facsimile of the first edition},
  author={Darwin, Charles},
  year={1964},
  publisher={Harvard University Press}
}

@article{miller2003cognitive,
  title={The cognitive revolution: a historical perspective},
  author={Miller, George A},
  journal={Trends in cognitive sciences},
  volume={7},
  number={3},
  pages={141--144},
  year={2003},
  publisher={Elsevier}
}

@incollection{newell2007computer,
  title={Computer science as empirical inquiry: Symbols and search},
  author={Newell, Allen and Simon, Herbert A},
  booktitle={ACM Turing award lectures},
  pages={1975},
  year={2007}
}

@inproceedings{hewitt-manning-2019-structural,
    title = "{A} Structural Probe for Finding Syntax in Word Representations",
    author = "Hewitt, John  and
      Manning, Christopher D.",
    editor = "Burstein, Jill  and
      Doran, Christy  and
      Solorio, Thamar",
    booktitle = "Proceedings of the 2019 Conference of the North {A}merican Chapter of the Association for Computational Linguistics: Human Language Technologies, Volume 1 (Long and Short Papers)",
    month = jun,
    year = "2019",
    address = "Minneapolis, Minnesota",
    publisher = "Association for Computational Linguistics",
    url = "https://aclanthology.org/N19-1419/",
    doi = "10.18653/v1/N19-1419",
    pages = "4129--4138"
}

@misc{htut2019attentionheadsberttrack,
      title={Do Attention Heads in {BERT} Track Syntactic Dependencies?}, 
      author={Phu Mon Htut and Jason Phang and Shikha Bordia and Samuel R. Bowman},
      year={2019},
      eprint={1911.12246},
      archivePrefix={arXiv},
      primaryClass={cs.CL},
      url={https://arxiv.org/abs/1911.12246}, 
}

@inproceedings{
marks2024geometrytruthemergentlinear,
title={The Geometry of Truth: Emergent Linear Structure in Large Language Model Representations of True/False Datasets},
author={Samuel Marks and Max Tegmark},
booktitle={First Conference on Language Modeling},
year={2024},
url={https://openreview.net/forum?id=aajyHYjjsk}
}

@inproceedings{
hong2025impliesbcircuitanalysis,
title={{A} Implies {B}: Circuit Analysis in {LLM}s for Propositional Logical Reasoning},
author={Guan Zhe Hong and Nishanth Dikkala and Enming Luo and Cyrus Rashtchian and Xin Wang and Rina Panigrahy},
booktitle={The Thirty-ninth Annual Conference on Neural Information Processing Systems},
year={2025},
url={https://openreview.net/forum?id=M0U8wUow8c}
}

@article{newell1980physical,
  title={Physical symbol systems},
  author={Newell, Allen},
  journal={Cognitive science},
  volume={4},
  number={2},
  pages={135--183},
  year={1980},
  publisher={Elsevier}
}

@article{fodor1988connectionism,
  title={Connectionism and cognitive architecture: A critical analysis},
  author={Fodor, Jerry A and Pylyshyn, Zenon W},
  journal={Cognition},
  volume={28},
  number={1-2},
  pages={3--71},
  year={1988},
  publisher={Elsevier}
}

@inproceedings{
burns2024discoveringlatentknowledgelanguage,
title={Discovering Latent Knowledge in Language Models Without Supervision},
author={Collin Burns and Haotian Ye and Dan Klein and Jacob Steinhardt},
booktitle={The Eleventh International Conference on Learning Representations },
year={2023},
url={https://openreview.net/forum?id=ETKGuby0hcs}
}

@misc{turner2024steeringlanguagemodelsactivation,
      title={Steering Language Models With Activation Engineering}, 
      author={Alexander Matt Turner and Lisa Thiergart and Gavin Leech and David Udell and Juan J. Vazquez and Ulisse Mini and Monte MacDiarmid},
      year={2024},
      eprint={2308.10248},
      archivePrefix={arXiv},
      primaryClass={cs.CL},
      url={https://arxiv.org/abs/2308.10248}, 
}

@inproceedings{clark2019doesbertlookat,
    title = "What Does {BERT} Look at? An Analysis of {BERT}{'}s Attention",
    author = "Clark, Kevin  and
      Khandelwal, Urvashi  and
      Levy, Omer  and
      Manning, Christopher D.",
    editor = "Linzen, Tal  and
      Chrupa{\l}a, Grzegorz  and
      Belinkov, Yonatan  and
      Hupkes, Dieuwke",
    booktitle = "Proceedings of the 2019 ACL Workshop BlackboxNLP: Analyzing and Interpreting Neural Networks for NLP",
    month = aug,
    year = "2019",
    address = "Florence, Italy",
    publisher = "Association for Computational Linguistics",
    url = "https://aclanthology.org/W19-4828/",
    doi = "10.18653/v1/W19-4828",
    pages = "276--286"
}

@misc{singh2024needsrightinductionhead,
      title={What needs to go right for an induction head? A mechanistic study of in-context learning circuits and their formation}, 
      author={Aaditya K. Singh and Ted Moskovitz and Felix Hill and Stephanie C. Y. Chan and Andrew M. Saxe},
      year={2024},
      eprint={2404.07129},
      archivePrefix={arXiv},
      primaryClass={cs.LG},
      url={https://arxiv.org/abs/2404.07129}, 
}

@inproceedings{geva2021transformerfeedforwardlayerskeyvalue,
    title = "Transformer Feed-Forward Layers Are Key-Value Memories",
    author = "Geva, Mor  and
      Schuster, Roei  and
      Berant, Jonathan  and
      Levy, Omer",
    editor = "Moens, Marie-Francine  and
      Huang, Xuanjing  and
      Specia, Lucia  and
      Yih, Scott Wen-tau",
    booktitle = "Proceedings of the 2021 Conference on Empirical Methods in Natural Language Processing",
    month = nov,
    year = "2021",
    address = "Online and Punta Cana, Dominican Republic",
    publisher = "Association for Computational Linguistics",
    url = "https://aclanthology.org/2021.emnlp-main.446/",
    doi = "10.18653/v1/2021.emnlp-main.446",
    pages = "5484--5495"
}

@article{cammarata2020thread:,
  author = {Cammarata, Nick and Carter, Shan and Goh, Gabriel and Olah, Chris and Petrov, Michael and Schubert, Ludwig and Voss, Chelsea and Egan, Ben and Lim, Swee Kiat},
  title = {Thread: Circuits},
  journal = {Distill},
  year = {2020},
  note = {https://distill.pub/2020/circuits},
  doi = {10.23915/distill.00024}
}

@inproceedings{
nanda2023progressmeasuresgrokkingmechanistic,
title={Progress measures for grokking via mechanistic interpretability},
author={Neel Nanda and Lawrence Chan and Tom Lieberum and Jess Smith and Jacob Steinhardt},
booktitle={The Eleventh International Conference on Learning Representations },
year={2023},
url={https://openreview.net/forum?id=9XFSbDPmdW}
}

@article{mccoy2026discover,
  title={The Emergent Symbolic Structure of Artificial Neural Networks},
  author={McCoy, R Thomas and Soulos, Paul and Linzen, Tal and Smolensky, Paul},
  journal={arXiv},
  year={2026}
}

@misc{Olah2022MechInterpVariables,
  title        = {Mechanistic Interpretability, Variables, and the Importance of Interpretable Bases},
  author       = {Chris Olah},
  year         = {2022},
  month        = jun,
  howpublished = {\url{https://www.transformer-circuits.pub/2022/mech-interp-essay}},
  note         = {Transformer Circuits Thread},
  urldate      = {2025-12-27}
}

@misc{vera2025embeddinggemmapowerfullightweighttext,
      title={EmbeddingGemma: Powerful and Lightweight Text Representations}, 
      author={Henrique Schechter Vera and Sahil Dua and Biao Zhang and Daniel Salz and Ryan Mullins and Sindhu Raghuram Panyam and Sara Smoot and Iftekhar Naim and Joe Zou and Feiyang Chen and Daniel Cer and Alice Lisak and Min Choi and Lucas Gonzalez and Omar Sanseviero and Glenn Cameron and Ian Ballantyne and Kat Black and Kaifeng Chen and Weiyi Wang and Zhe Li and Gus Martins and Jinhyuk Lee and Mark Sherwood and Juyeong Ji and Renjie Wu and Jingxiao Zheng and Jyotinder Singh and Abheesht Sharma and Divyashree Sreepathihalli and Aashi Jain and Adham Elarabawy and AJ Co and Andreas Doumanoglou and Babak Samari and Ben Hora and Brian Potetz and Dahun Kim and Enrique Alfonseca and Fedor Moiseev and Feng Han and Frank Palma Gomez and Gustavo Hernández Ábrego and Hesen Zhang and Hui Hui and Jay Han and Karan Gill and Ke Chen and Koert Chen and Madhuri Shanbhogue and Michael Boratko and Paul Suganthan and Sai Meher Karthik Duddu and Sandeep Mariserla and Setareh Ariafar and Shanfeng Zhang and Shijie Zhang and Simon Baumgartner and Sonam Goenka and Steve Qiu and Tanmaya Dabral and Trevor Walker and Vikram Rao and Waleed Khawaja and Wenlei Zhou and Xiaoqi Ren and Ye Xia and Yichang Chen and Yi-Ting Chen and Zhe Dong and Zhongli Ding and Francesco Visin and Gaël Liu and Jiageng Zhang and Kathleen Kenealy and Michelle Casbon and Ravin Kumar and Thomas Mesnard and Zach Gleicher and Cormac Brick and Olivier Lacombe and Adam Roberts and Qin Yin and Yunhsuan Sung and Raphael Hoffmann and Tris Warkentin and Armand Joulin and Tom Duerig and Mojtaba Seyedhosseini},
      year={2025},
      eprint={2509.20354},
      archivePrefix={arXiv},
      primaryClass={cs.CL},
      url={https://arxiv.org/abs/2509.20354}, 
}

@inproceedings{warner2024smarterbetterfasterlonger,
    title = "Smarter, Better, Faster, Longer: A Modern Bidirectional Encoder for Fast, Memory Efficient, and Long Context Finetuning and Inference",
    author = {Warner, Benjamin  and
      Chaffin, Antoine  and
      Clavi{\'e}, Benjamin  and
      Weller, Orion  and
      Hallstr{\"o}m, Oskar  and
      Taghadouini, Said  and
      Gallagher, Alexis  and
      Biswas, Raja  and
      Ladhak, Faisal  and
      Aarsen, Tom  and
      Adams, Griffin Thomas  and
      Howard, Jeremy  and
      Poli, Iacopo},
    editor = "Che, Wanxiang  and
      Nabende, Joyce  and
      Shutova, Ekaterina  and
      Pilehvar, Mohammad Taher",
    booktitle = "Proceedings of the 63rd Annual Meeting of the Association for Computational Linguistics (Volume 1: Long Papers)",
    month = jul,
    year = "2025",
    address = "Vienna, Austria",
    publisher = "Association for Computational Linguistics",
    url = "https://aclanthology.org/2025.acl-long.127/",
    doi = "10.18653/v1/2025.acl-long.127",
    pages = "2526--2547",
    ISBN = "979-8-89176-251-0"
}

@misc{zhang2025qwen3embeddingadvancingtext,
      title={Qwen3 Embedding: Advancing Text Embedding and Reranking Through Foundation Models}, 
      author={Yanzhao Zhang and Mingxin Li and Dingkun Long and Xin Zhang and Huan Lin and Baosong Yang and Pengjun Xie and An Yang and Dayiheng Liu and Junyang Lin and Fei Huang and Jingren Zhou},
      year={2025},
      eprint={2506.05176},
      archivePrefix={arXiv},
      primaryClass={cs.CL},
      url={https://arxiv.org/abs/2506.05176}, 
}

@inproceedings{reimers-2019-sentence-bert,
  title = "Sentence-BERT: Sentence Embeddings using Siamese BERT-Networks",
  author = "Reimers, Nils and Gurevych, Iryna",
  booktitle = "Proceedings of the 2019 Conference on Empirical Methods in Natural Language Processing",
  month = "11",
  year = "2019",
  publisher = "Association for Computational Linguistics",
  url = "https://arxiv.org/abs/1908.10084",
}

@inproceedings{
gao2024scalingevaluatingsparseautoencoders,
title={Scaling and evaluating sparse autoencoders},
author={Leo Gao and Tom Dupre la Tour and Henk Tillman and Gabriel Goh and Rajan Troll and Alec Radford and Ilya Sutskever and Jan Leike and Jeffrey Wu},
booktitle={The Thirteenth International Conference on Learning Representations},
year={2025},
url={https://openreview.net/forum?id=tcsZt9ZNKD}
}

@article{templeton2024scaling,
   title={Scaling Monosemanticity: Extracting Interpretable Features from {Claude 3 Sonnet}},
   author={Templeton, Adly and Conerly, Tom and Marcus, Jonathan and Lindsey, Jack and Bricken, Trenton and Chen, Brian and Pearce, Adam and Citro, Craig and Ameisen, Emmanuel and Jones, Andy and Cunningham, Hoagy and Turner, Nicholas L and McDougall, Callum and MacDiarmid, Monte and Freeman, C. Daniel and Sumers, Theodore R. and Rees, Edward and Batson, Joshua and Jermyn, Adam and Carter, Shan and Olah, Chris and Henighan, Tom},
   year={2024},
   journal={Transformer Circuits Thread},
   url={https://transformer-circuits.pub/2024/scaling-monosemanticity/index.html}
}

@article{bricken2023monosemanticity,
   title={Towards Monosemanticity: Decomposing Language Models With Dictionary Learning},
   author={Bricken, Trenton and Templeton, Adly and Batson, Joshua and Chen, Brian and Jermyn, Adam and Conerly, Tom and Turner, Nick and Anil, Cem and Denison, Carson and Askell, Amanda and Lasenby, Robert and Wu, Yifan and Kravec, Shauna and Schiefer, Nicholas and Maxwell, Tim and Joseph, Nicholas and Hatfield-Dodds, Zac and Tamkin, Alex and Nguyen, Karina and McLean, Brayden and Burke, Josiah E and Hume, Tristan and Carter, Shan and Henighan, Tom and Olah, Christopher},
   year={2023},
   journal={Transformer Circuits Thread},
   note={https://transformer-circuits.pub/2023/monosemantic-features/index.html}
}

@article{Kleyko_2022a,
   title={Vector Symbolic Architectures as a Computing Framework for Emerging Hardware},
   volume={110},
   ISSN={1558-2256},
   url={http://dx.doi.org/10.1109/JPROC.2022.3209104},
   DOI={10.1109/jproc.2022.3209104},
   number={10},
   journal={Proceedings of the IEEE},
   publisher={Institute of Electrical and Electronics Engineers (IEEE)},
   author={Kleyko, Denis and Davies, Mike and Frady, Edward Paxon and Kanerva, Pentti and Kent, Spencer J. and Olshausen, Bruno A. and Osipov, Evgeny and Rabaey, Jan M. and Rachkovskij, Dmitri A. and Rahimi, Abbas and Sommer, Friedrich T.},
   year={2022},
   month=oct, pages={1538–1571} }

@article{Kleyko_2022b,
   title={A Survey on Hyperdimensional Computing aka Vector Symbolic Architectures, Part I: Models and Data Transformations},
   volume={55},
   ISSN={1557-7341},
   url={http://dx.doi.org/10.1145/3538531},
   DOI={10.1145/3538531},
   number={6},
   journal={ACM Computing Surveys},
   publisher={Association for Computing Machinery (ACM)},
   author={Kleyko, Denis and Rachkovskij, Dmitri A. and Osipov, Evgeny and Rahimi, Abbas},
   year={2022},
   month=dec, pages={1–40} }

@misc{olmo20252olmo2furious,
      title={2 {OLMo} 2 {Furious}}, 
      author={Team OLMo and Pete Walsh and Luca Soldaini and Dirk Groeneveld and Kyle Lo and Shane Arora and Akshita Bhagia and Yuling Gu and Shengyi Huang and Matt Jordan and Nathan Lambert and Dustin Schwenk and Oyvind Tafjord and Taira Anderson and David Atkinson and Faeze Brahman and Christopher Clark and Pradeep Dasigi and Nouha Dziri and Allyson Ettinger and Michal Guerquin and David Heineman and Hamish Ivison and Pang Wei Koh and Jiacheng Liu and Saumya Malik and William Merrill and Lester James V. Miranda and Jacob Morrison and Tyler Murray and Crystal Nam and Jake Poznanski and Valentina Pyatkin and Aman Rangapur and Michael Schmitz and Sam Skjonsberg and David Wadden and Christopher Wilhelm and Michael Wilson and Luke Zettlemoyer and Ali Farhadi and Noah A. Smith and Hannaneh Hajishirzi},
      year={2025},
      eprint={2501.00656},
      archivePrefix={arXiv},
      primaryClass={cs.CL},
      url={https://arxiv.org/abs/2501.00656}, 
}

@inproceedings{linzen-2016-issues,
    title = "Issues in evaluating semantic spaces using word analogies",
    author = "Linzen, Tal",
    booktitle = "Proceedings of the 1st Workshop on Evaluating Vector-Space Representations for {NLP}",
    month = aug,
    year = "2016",
    address = "Berlin, Germany",
    publisher = "Association for Computational Linguistics",
    url = "https://aclanthology.org/W16-2503/",
    doi = "10.18653/v1/W16-2503",
    pages = "13--18"
}

@article{hupkes2018visualisationdiagnosticclassifiersreveal,
author = {Hupkes, Dieuwke and Veldhoen, Sara and Zuidema, Willem},
title = {Visualisation and ‘diagnostic classifiers’ reveal how recurrent and recursive neural networks process hierarchical structure},
year = {2018},
issue_date = {January 2018},
publisher = {AI Access Foundation},
address = {El Segundo, CA, USA},
volume = {61},
number = {1},
issn = {1076-9757},
journal = {J. Artif. Int. Res.},
month = jan,
pages = {907–926},
numpages = {20}
}

@inproceedings{ettinger-etal-2016-probing,
    title = "Probing for semantic evidence of composition by means of simple classification tasks",
    author = "Ettinger, Allyson  and
      Elgohary, Ahmed  and
      Resnik, Philip",
    booktitle = "Proceedings of the 1st Workshop on Evaluating Vector-Space Representations for {NLP}",
    month = aug,
    year = "2016",
    address = "Berlin, Germany",
    publisher = "Association for Computational Linguistics",
    url = "https://aclanthology.org/W16-2524/",
    doi = "10.18653/v1/W16-2524",
    pages = "134--139"
}

@inproceedings{csordás2024recurrentneuralnetworkslearn,
    title = "Recurrent Neural Networks Learn to Store and Generate Sequences using Non-Linear Representations",
    author = "Csord{\'a}s, R{\'o}bert  and
      Potts, Christopher  and
      Manning, Christopher D  and
      Geiger, Atticus",
    editor = "Belinkov, Yonatan  and
      Kim, Najoung  and
      Jumelet, Jaap  and
      Mohebbi, Hosein  and
      Mueller, Aaron  and
      Chen, Hanjie",
    booktitle = "Proceedings of the 7th BlackboxNLP Workshop: Analyzing and Interpreting Neural Networks for NLP",
    month = nov,
    year = "2024",
    address = "Miami, Florida, US",
    publisher = "Association for Computational Linguistics",
    url = "https://aclanthology.org/2024.blackboxnlp-1.17/",
    doi = "10.18653/v1/2024.blackboxnlp-1.17",
    pages = "248--262"
}

@inproceedings{
paulo2025sparseautoencoderstraineddata,
title={Sparse Autoencoders Trained on the Same Data Learn Different Features},
author={Gon{\c{c}}alo Paulo and Nora Belrose},
booktitle={The Fourteenth International Conference on Learning Representations},
year={2026},
url={https://openreview.net/forum?id=EjInprGpk9}
}

@misc{tian2025measuringsparseautoencoderfeature,
      title={Measuring Sparse Autoencoder Feature Sensitivity}, 
      author={Claire Tian and Katherine Tian and Nathan Hu},
      year={2025},
      eprint={2509.23717},
      archivePrefix={arXiv},
      primaryClass={cs.AI},
      url={https://arxiv.org/abs/2509.23717}, 
}

@inproceedings{
makelov2025towards,
title={Towards Principled Evaluations of Sparse Autoencoders for Interpretability and Control},
author={Aleksandar Makelov and Georg Lange and Neel Nanda},
booktitle={The Thirteenth International Conference on Learning Representations},
year={2025},
url={https://openreview.net/forum?id=1Njl73JKjB}
}

@inproceedings{
karvonen2024measuringprogressdictionarylearning,
title={Measuring Progress in Dictionary Learning for Language Model Interpretability with Board Game Models},
author={Adam Karvonen and Benjamin Wright and Can Rager and Rico Angell and Jannik Brinkmann and Logan Riggs Smith and Claudio Mayrink Verdun and David Bau and Samuel Marks},
booktitle={ICML 2024 Workshop on Mechanistic Interpretability},
year={2024},
url={https://openreview.net/forum?id=qzsDKwGJyB}
}

@inproceedings{
geiger2021causal,
title={Causal Abstractions of Neural Networks},
author={Atticus Geiger and Hanson Lu and Thomas F Icard and Christopher Potts},
booktitle={Advances in Neural Information Processing Systems},
editor={A. Beygelzimer and Y. Dauphin and P. Liang and J. Wortman Vaughan},
year={2021},
url={https://openreview.net/forum?id=RmuXDtjDhG}
}

@misc{goldowskydill2023localizingmodelbehaviorpath,
      title={Localizing Model Behavior with Path Patching}, 
      author={Nicholas Goldowsky-Dill and Chris MacLeod and Lucas Sato and Aryaman Arora},
      year={2023},
      eprint={2304.05969},
      archivePrefix={arXiv},
      primaryClass={cs.LG},
      url={https://arxiv.org/abs/2304.05969}, 
}

@inproceedings{cho-etal-2014-learning,
    title = "Learning Phrase Representations using {RNN} Encoder{--}Decoder for Statistical Machine Translation",
    author = {Cho, Kyunghyun  and
      van Merri{\"e}nboer, Bart  and
      Gulcehre, Caglar  and
      Bahdanau, Dzmitry  and
      Bougares, Fethi  and
      Schwenk, Holger  and
      Bengio, Yoshua},
    editor = "Moschitti, Alessandro  and
      Pang, Bo  and
      Daelemans, Walter",
    booktitle = "Proceedings of the 2014 Conference on Empirical Methods in Natural Language Processing ({EMNLP})",
    month = oct,
    year = "2014",
    address = "Doha, Qatar",
    publisher = "Association for Computational Linguistics",
    url = "https://aclanthology.org/D14-1179/",
    doi = "10.3115/v1/D14-1179",
    pages = "1724--1734"
}

@ARTICLE{hochreiter1997lstm,
  author={Hochreiter, Sepp and Schmidhuber, Jürgen},
  journal={Neural Computation}, 
  title={Long Short-Term Memory}, 
  year={1997},
  volume={9},
  number={8},
  pages={1735-1780},
  doi={10.1162/neco.1997.9.8.1735}}

@article{ELMAN1990179,
	author = {Jeffrey L. Elman},
	doi = {https://doi.org/10.1016/0364-0213(90)90002-E},
	issn = {0364-0213},
	journal = {Cognitive Science},
	number = {2},
	pages = {179-211},
	title = {Finding structure in time},
	url = {https://www.sciencedirect.com/science/article/pii/036402139090002E},
	volume = {14},
	year = {1990}}

@misc{makhzani2014ksparseautoencoders,
      title={k-Sparse Autoencoders}, 
      author={Alireza Makhzani and Brendan Frey},
      year={2014},
      eprint={1312.5663},
      archivePrefix={arXiv},
      primaryClass={cs.LG},
      url={https://arxiv.org/abs/1312.5663}, 
}

@inproceedings{
rajamanoharan2024improvingdictionarylearninggated,
title={Improving Sparse Decomposition of Language Model Activations with Gated Sparse Autoencoders},
author={Senthooran Rajamanoharan and Arthur Conmy and Lewis Smith and Tom Lieberum and Vikrant Varma and Janos Kramar and Rohin Shah and Neel Nanda},
booktitle={The Thirty-eighth Annual Conference on Neural Information Processing Systems},
year={2024},
url={https://openreview.net/forum?id=zLBlin2zvW}
}

@InProceedings{pmlr-v236-geiger24a,
  title = 	 {Finding Alignments Between Interpretable Causal Variables and Distributed Neural Representations},
  author =       {Geiger, Atticus and Wu, Zhengxuan and Potts, Christopher and Icard, Thomas and Goodman, Noah},
  booktitle = 	 {Proceedings of the Third Conference on Causal Learning and Reasoning},
  pages = 	 {160--187},
  year = 	 {2024},
  editor = 	 {Locatello, Francesco and Didelez, Vanessa},
  volume = 	 {236},
  series = 	 {Proceedings of Machine Learning Research},
  month = 	 {01--03 Apr},
  publisher =    {PMLR},
  url = 	 {https://proceedings.mlr.press/v236/geiger24a.html}
}

@inproceedings{hendel-etal-2023-context,
    title = "In-Context Learning Creates Task Vectors",
    author = "Hendel, Roee  and
      Geva, Mor  and
      Globerson, Amir",
    editor = "Bouamor, Houda  and
      Pino, Juan  and
      Bali, Kalika",
    booktitle = "Findings of the Association for Computational Linguistics: EMNLP 2023",
    month = dec,
    year = "2023",
    address = "Singapore",
    publisher = "Association for Computational Linguistics",
    url = "https://aclanthology.org/2023.findings-emnlp.624/",
    doi = "10.18653/v1/2023.findings-emnlp.624",
    pages = "9318--9333"
}

@inproceedings{
todd2024functionvectorslargelanguage,
title={Function Vectors in Large Language Models},
author={Eric Todd and Millicent Li and Arnab Sen Sharma and Aaron Mueller and Byron C Wallace and David Bau},
booktitle={The Twelfth International Conference on Learning Representations},
year={2024},
url={https://openreview.net/forum?id=AwyxtyMwaG}
}

@inproceedings{
pres2024reliableevaluationbehaviorsteering,
title={Towards Reliable Evaluation of Behavior Steering Interventions in {LLM}s},
author={Itamar Pres and Laura Ruis and Ekdeep Singh Lubana and David Krueger},
booktitle={MINT: Foundation Model Interventions},
year={2024},
url={https://openreview.net/forum?id=7xJcX2gbm9}
}

@article{
bau2017networkdissection,
author = {David Bau  and Jun-Yan Zhu  and Hendrik Strobelt  and Agata Lapedriza  and Bolei Zhou  and Antonio Torralba },
title = {Understanding the role of individual units in a deep neural network},
journal = {Proceedings of the National Academy of Sciences},
volume = {117},
number = {48},
pages = {30071-30078},
year = {2020},
doi = {10.1073/pnas.1907375117},
URL = {https://www.pnas.org/doi/abs/10.1073/pnas.1907375117},
eprint = {https://www.pnas.org/doi/pdf/10.1073/pnas.1907375117}}

@misc{kramár2024atpefficientscalablemethod,
      title={AtP*: An efficient and scalable method for localizing {LLM} behaviour to components}, 
      author={János Kramár and Tom Lieberum and Rohin Shah and Neel Nanda},
      year={2024},
      eprint={2403.00745},
      archivePrefix={arXiv},
      primaryClass={cs.LG},
      url={https://arxiv.org/abs/2403.00745}, 
}

@misc{yang2025qwen3technicalreport,
      title={Qwen3 Technical Report}, 
      author={An Yang and Anfeng Li and Baosong Yang and Beichen Zhang and Binyuan Hui and Bo Zheng and Bowen Yu and Chang Gao and Chengen Huang and Chenxu Lv and Chujie Zheng and Dayiheng Liu and Fan Zhou and Fei Huang and Feng Hu and Hao Ge and Haoran Wei and Huan Lin and Jialong Tang and Jian Yang and Jianhong Tu and Jianwei Zhang and Jianxin Yang and Jiaxi Yang and Jing Zhou and Jingren Zhou and Junyang Lin and Kai Dang and Keqin Bao and Kexin Yang and Le Yu and Lianghao Deng and Mei Li and Mingfeng Xue and Mingze Li and Pei Zhang and Peng Wang and Qin Zhu and Rui Men and Ruize Gao and Shixuan Liu and Shuang Luo and Tianhao Li and Tianyi Tang and Wenbiao Yin and Xingzhang Ren and Xinyu Wang and Xinyu Zhang and Xuancheng Ren and Yang Fan and Yang Su and Yichang Zhang and Yinger Zhang and Yu Wan and Yuqiong Liu and Zekun Wang and Zeyu Cui and Zhenru Zhang and Zhipeng Zhou and Zihan Qiu},
      year={2025},
      eprint={2505.09388},
      archivePrefix={arXiv},
      primaryClass={cs.CL},
      url={https://arxiv.org/abs/2505.09388}, 
}

@inproceedings{
engels2025languagemodelfeaturesonedimensionally,
title={Not All Language Model Features Are One-Dimensionally Linear},
author={Joshua Engels and Eric J Michaud and Isaac Liao and Wes Gurnee and Max Tegmark},
booktitle={The Thirteenth International Conference on Learning Representations},
year={2025},
url={https://openreview.net/forum?id=d63a4AM4hb}
}

@inproceedings{razzhigaev2025llmmicroscopeuncoveringhiddenrole,
    title = "{LLM}-Microscope: Uncovering the Hidden Role of Punctuation in Context Memory of Transformers",
    author = "Razzhigaev, Anton  and
      Mikhalchuk, Matvey  and
      Rahmatullaev, Temurbek  and
      Goncharova, Elizaveta  and
      Druzhinina, Polina  and
      Oseledets, Ivan  and
      Kuznetsov, Andrey",
    editor = "Chiruzzo, Luis  and
      Ritter, Alan  and
      Wang, Lu",
    booktitle = "Findings of the Association for Computational Linguistics: NAACL 2025",
    month = apr,
    year = "2025",
    address = "Albuquerque, New Mexico",
    publisher = "Association for Computational Linguistics",
    url = "https://aclanthology.org/2025.findings-naacl.432/",
    doi = "10.18653/v1/2025.findings-naacl.432",
    pages = "7772--7779",
    ISBN = "979-8-89176-195-7"
}

@misc{openai2025gptoss120bgptoss20bmodel,
      title={gpt-oss-120b \& gpt-oss-20b Model Card}, 
      author={OpenAI and : and Sandhini Agarwal and Lama Ahmad and Jason Ai and Sam Altman and Andy Applebaum and Edwin Arbus and Rahul K. Arora and Yu Bai and Bowen Baker and Haiming Bao and Boaz Barak and Ally Bennett and Tyler Bertao and Nivedita Brett and Eugene Brevdo and Greg Brockman and Sebastien Bubeck and Che Chang and Kai Chen and Mark Chen and Enoch Cheung and Aidan Clark and Dan Cook and Marat Dukhan and Casey Dvorak and Kevin Fives and Vlad Fomenko and Timur Garipov and Kristian Georgiev and Mia Glaese and Tarun Gogineni and Adam Goucher and Lukas Gross and Katia Gil Guzman and John Hallman and Jackie Hehir and Johannes Heidecke and Alec Helyar and Haitang Hu and Romain Huet and Jacob Huh and Saachi Jain and Zach Johnson and Chris Koch and Irina Kofman and Dominik Kundel and Jason Kwon and Volodymyr Kyrylov and Elaine Ya Le and Guillaume Leclerc and James Park Lennon and Scott Lessans and Mario Lezcano-Casado and Yuanzhi Li and Zhuohan Li and Ji Lin and Jordan Liss and Lily and Liu and Jiancheng Liu and Kevin Lu and Chris Lu and Zoran Martinovic and Lindsay McCallum and Josh McGrath and Scott McKinney and Aidan McLaughlin and Song Mei and Steve Mostovoy and Tong Mu and Gideon Myles and Alexander Neitz and Alex Nichol and Jakub Pachocki and Alex Paino and Dana Palmie and Ashley Pantuliano and Giambattista Parascandolo and Jongsoo Park and Leher Pathak and Carolina Paz and Ludovic Peran and Dmitry Pimenov and Michelle Pokrass and Elizabeth Proehl and Huida Qiu and Gaby Raila and Filippo Raso and Hongyu Ren and Kimmy Richardson and David Robinson and Bob Rotsted and Hadi Salman and Suvansh Sanjeev and Max Schwarzer and D. Sculley and Harshit Sikchi and Kendal Simon and Karan Singhal and Yang Song and Dane Stuckey and Zhiqing Sun and Philippe Tillet and Sam Toizer and Foivos Tsimpourlas and Nikhil Vyas and Eric Wallace and Xin Wang and Miles Wang and Olivia Watkins and Kevin Weil and Amy Wendling and Kevin Whinnery and Cedric Whitney and Hannah Wong and Lin Yang and Yu Yang and Michihiro Yasunaga and Kristen Ying and Wojciech Zaremba and Wenting Zhan and Cyril Zhang and Brian Zhang and Eddie Zhang and Shengjia Zhao},
      year={2025},
      eprint={2508.10925},
      archivePrefix={arXiv},
      primaryClass={cs.CL},
      url={https://arxiv.org/abs/2508.10925}, 
}

@inproceedings{
fiottokaufman2024nnsightndifdemocratizingaccess,
title={{NN}sight and {NDIF}: Democratizing Access to Open-Weight Foundation Model Internals},
author={Jaden Fried Fiotto-Kaufman and Alexander Russell Loftus and Eric Todd and Jannik Brinkmann and Koyena Pal and Dmitrii Troitskii and Michael Ripa and Adam Belfki and Can Rager and Caden Juang and Aaron Mueller and Samuel Marks and Arnab Sen Sharma and Francesca Lucchetti and Nikhil Prakash and Carla E. Brodley and Arjun Guha and Jonathan Bell and Byron C Wallace and David Bau},
booktitle={The Thirteenth International Conference on Learning Representations},
year={2025},
url={https://openreview.net/forum?id=MxbEiFRf39}
}


\newpage
\appendix
\section{Overview of appendices}

These appendices provide mathematical derivations, experimental details, and extended results that complement the main text. The appendices start with tables listing notations and abbreviations used throughout the paper. Then, separate sections detail the synthetic sequence and SVO sentence datasets, then list model and training hyperparameters, followed by the evaluation procedures for linear probing, analogies, sparse autoencoders, and activation patching.

\section{Finding unbinding vectors}
\label{ap:tpr-unbinding}

Given a set of role vectors $\{r_1, r_2, \dots, r_n\}$ and the role matrix $R$ where each column is a role vector, \emph{unbinding} vectors are vectors $u_i$ such that for all $u$ and $r$'s, $u_i \cdot r_j = \delta_{ij}$. Informally, the dot product between the unbinding vector and a role vector should be 1 if and only if they correspond, and should be 0 otherwise. Finding such unbinding vectors would allow for lossless information retrieval from TPRs, as Section~\ref{sec:vsa-tprs} shows. 

If we denote the unbinding matrix $U$, where each column is an unbinding vector corresponding to the role vector in the same column, it is not difficult to see that we have $U^\top R = I$ if and only if $R$ is full-rank. As role vectors are real-valued, then as long as $|\{R\}| \leq d_r$, i.e., the number of roles is less than or equal to the dimension of role vectors, $R$ is very likely full-rank.\footnote{To make a hand-wavy argument, $R$ will be full rank with probability 1 assuming any reasonable family of continuous distribution for $r$.} Then, finding the unbinding vectors simply requires computing the inverse of the role matrix. 

For this reason, we limit our role dimensions choices to those larger than the number of roles in all of our experiments.\footnote{TPRs also handle approximate unbinding as introduced in \cite{Smolensky1990TPRs}, which allows for packing significantly more role vectors within a fixed high-dimensional space under a certain unbinding error limit. Here we circumvent the need to consider approximate unbinding as it is outside the scope of the project.} We also compute the regularized pseudoinverse for the unbinding vectors as gradient descent finds noisy approximations of role vectors. We find that naively using the Tikhonov-regularized pseudoinverse with $\lambda = 0.1$ works well for all our experiments.

\section{Recovering a TPR from neural network hidden state}

A shared component between constructing linear probes (Appendix~\ref{ap:probe-tpe-derivation}) and SAEs (Appendx~\ref{ap:sae-tpe-derivation}), or more generally, any application of TPR unbinding, is that they must first recover a TPR from the neural network hidden states. 

This step is non trivial. To give an example, the RNN we trained TPEs to approximate (Section~\ref{sec:setup-seqs}) has a hidden state size of 256, while the TPR binding has $64 \times 64 = 4096$ elements when flattened. As such, the projection weights $W$ form a down projection for most of the Tensor Product Encoder setups, and recovering such a TPR would require computing an up-projection that entails an ill-conditioned pseudoinverse.

One option, of course, is to constrain the dimensions of fillers and role such that the TPR always has fewer elements than the activation it is trained to approximate. Empirically, however, we find that this is not necessary or helpful. Instead, we find that merely using the pseudoinverse of $W$ (in an appropriately regularized form, as described below) is often possible, possibly because the flattened TPR does not span its full vector space. We offer no principled explanations for why this is the case, but provide a few observations from initial experiments:

\begin{enumerate}
    \item If one trains a sequence-to-sequence model that reverses sequence with a hidden size of 400, and then trains TPEs with filler and role dimension of 20 to approximate it, computing the (pseudo)inverse of the output projection layer gives an inverse transform that is able to recover a TPR from its linearly transformed variant, but not the NN hidden states it is trained to approximate, even though the two are closely correlated and the linear projection layer is in principle invertible.
    \item If one computes all of the singular values of the projection layer weight in the above-mentioned TPE, there will be a large number of small singular values that are approximately 0, suggesting that the projection layer is not in fact full rank. This finding is, in hindsight, unsurprising, as specifying a hyperparameter, the dimensionality of filler or role embeddings, bears no weight on the representational structure of the sequence-to-sequence network and the subspace it spans.
\end{enumerate}

Both the observations above and the up-projection generally required are similar to finding the linear best fit of an ill-posed regression problem. We thus apply the solution, applying a regularization technique called the Tikhonov regularization:
\begin{equation}
W^+_\lambda \;=\; (W^\top W + \lambda I)^{-1} W^\top,
\end{equation}
Where $\lambda \ge 0$. Intuitively, setting a larger $\lambda$ ``dampens'' the effect of small singular values on the pseudoinverse. We find that setting this regularization parameter correctly is crucial to retrieving an approximate TPR from a neural network hidden state. In our empirical experiments, the output layer regularization $\lambda$ is chosen with ternary search with respect to a batch of neural network activations and their filler-role encoding: we start with a range of $[10^{-12}, 10^{12}]$. At each iteration, we compute the pseudoinverse with the candidate regularization parameter, and then apply it on a batch of 128 neural network activations to obtain approximate TPRs and compute its MSE loss with the TPRs encoded with filler-role pairs. The algorithm repeats until it finds a $\pm10\%$ range of possible regularization parameters. The pseudocode for the full algorithm is in Algorithm~\ref{alg:ternary}.

\begin{algorithm}
\caption{Ternary search for regularization using a batch objective}
\label{alg:ternary}
\begin{lstlisting}
# inputs:
#   batch_hidden: the neural network hidden states of a batch of data
#   batch_filler_ids: the filler ids for the batch of data
#   batch_role_ids: the role ids for the batch of data
#   objective_fn: callable mapping log_lambda -> scalar loss
#   log_lo, log_hi: search bounds in log10(lambda)
#   tol: relative precision
# obtains:
#   best_lambda: lambda minimizing objective_fn

def ternary_search_log_lambda(objective_fn, log_lo, log_hi, tol):
    precision = log10(1 + tol)

    while (log_hi - log_lo) > precision:
        left_third  = (2 * log_lo + log_hi) / 3
        right_third = (log_lo + 2 * log_hi) / 3

        loss_left  = objective_fn(left_third)
        loss_right = objective_fn(right_third)

        if loss_left < loss_right:   # minimize
            log_hi = right_third
        else:
            log_lo = left_third

    best_log = 0.5 * (log_lo + log_hi)
    best_loss = objective_fn(best_log)
    return best_log, best_loss, (log_lo, log_hi)


# --- batch objective used for output-layer inversion ---
# inputs:
#   batch_hidden: [B, D] hidden states to invert
#   batch_filler_ids: [B, N], batch_role_ids: [B, N]
#   model: provides output-layer inverse and ground-truth TPR

def batch_objective(log_lambda):
    W_inv = output_layer_inverse(model, l2_lambda=10**log_lambda)
    recovered_tpr = linear(batch_hidden, W_inv)
    gt_tpr = model_tpr(batch_filler_ids, batch_role_ids)
    return mse(recovered_tpr, gt_tpr)


# call:
best_log, best_loss, bracket = ternary_search_log_lambda(
    objective_fn=batch_objective,
    log_lo=LOG_LO,
    log_hi=LOG_HI,
    tol=TOL,
)
best_lambda = 10**best_log
\end{lstlisting}
\end{algorithm}

\newpage
\section{Constructing linear probes from TPRs}
\label{ap:probe-tpe-derivation}

\paragraph{Outline} The derivation takes three main steps:

\begin{enumerate}
    \item recover $\mathrm{vec}(E)$ from $h$.
    \item unbind the recovered TPR with $u_j$
    \item project the unbinding result, which is a predicted filler embedding, onto filler vocabulary
\end{enumerate}

We provide a complete derivation of the analytic probe used in Section~\ref{sec:linear-probing}. We start with the standard TPR setup:

\paragraph{TPR setup} Let fillers $f_i \in \mathbb{R}^{d_f}$ and roles $r_i \in \mathbb{R}^{d_r}$. The (unflattened/unprojected) TPR matrix is
\begin{equation}
E \;=\; \sum_{i=1}^n f_i r_i^\top \;\in\; \mathbb{R}^{d_f \times d_r}.
\end{equation}
A Tensor Product Encoder maps $E$ to a hidden state $h \in \mathbb{R}^{d_h}$ by
\begin{equation}
h \;=\; W\,\text{vec}(E) + b,
\end{equation}
with projection $W \in \mathbb{R}^{d_h \times (d_f d_r)}$ and bias $b \in \mathbb{R}^{d_h}$. Let $u_j$ be the unbinding vector for role $r_j$, so that $r_i^\top u_j \approx \delta_{ij}$.

\paragraph{Lemma 1 (unbinding identity).} For any $E \in \mathbb{R}^{d_f \times d_r}$ and $u \in \mathbb{R}^{d_r}$,
\begin{equation}
E\,u \;=\; (u^\top \otimes I_{d_f})\,\text{vec}(E). \label{eq:unbinding-vec-identity}
\end{equation}
\emph{Proof.} This is the standard Kronecker/vec identity $(x^\top \otimes I_m)\,\text{vec}(A)=A x$.

\paragraph{Lemma 2 (recovering the filler from $h$).} Let $W^+$ denote the pseudoinverse of $W$, such that $W^+W \approx I$. Then the filler $f_j$ bound to role $r_j$ can be (approximately) recovered via the following:

$$(u_j^\top \otimes I_{d_f})\,W^+(h-b).$$

\emph{Proof.} First, the flattened TPR $\text{vec}(E)$ can be recovered via the following
\begin{align}
    h &= W(\text{vec}(E)) + b \\
    \hat{\text{vec}(E)} &:= W^+(h-b) = W^+((W(\text{vec}(E)) + b) - b) \approx \text{vec}(E)
\end{align}
Plugging $\hat{\text{vec}(E)}$ into Equation~\ref{eq:unbinding-vec-identity} gives
\begin{align}
    \hat E u \;&=\; (u_j^\top \otimes I_{d_f})\,W^+(h-b),
\end{align}
where the left hand side is the TPR unbinding operation (Equation~\ref{eq:tpr-unbind}). Thus,
\begin{equation}
\hat{f}_j \;=\; (u_j^\top \otimes I_{d_f})\,W^+(h-b),
\end{equation}
which approximates the filler bound to role $r_j$. Note that when the inverse $W^+$ is the exact inverse ($W^+W = I$), the filler is exactly recovered instead of approximately recovered.

\paragraph{Classification over the filler vocabulary} Let $F \in \mathbb{R}^{|\mathcal{F}| \times d_f}$ be the matrix of fillers used for classification (rows are filler vectors, optionally normalized). The logits for role $r_j$ are
\begin{equation}
\label{eq:tpe-probe}
\text{logits}(h) \;=\; F\,\hat{f}_j \;=\; F (u_j^\top \otimes I_{d_f}) W^+(h-b).
\end{equation}
Thus the analytic probe is linear in $h$ with parameters
\begin{align}
W_{\text{probe}} &= F (u_j^\top \otimes I_{d_f}) W^+, &
b_{\text{probe}} &= -W_{\text{probe}}\, b.
\end{align}
This yields a fully specified linear probe with no additional training.

\section{Constructing SAEs from TPRs}
\label{ap:sae-tpe-derivation}

\paragraph{Outline} We consider deriving individual feature vectors, which we then compose into an SAE by constructing feature vectors for all possible concepts. The derivation takes 3 main steps, in a structure analogous to linear probe construction:

\begin{enumerate}
    \item express a desired binding score as a linear function of $\text{vec}(E)$.
    \item substitute the TPE inverse to obtain a linear detector in $h$.
    \item embed the detector as an SAE feature via $(W_{\text{enc}}, b_{\text{enc}})$ and set a compatible decoder.
\end{enumerate}

\paragraph{Setup.} Let $h = W\,\text{vec}(E)+b$ be a linearly transformed TPR, with $E=\sum_i f_i r_i^\top$. An SAE with hidden features $z \in \mathbb{R}^m$ has the form
\begin{align}
z &= \text{ReLU}(W_{\text{enc}} h + b_{\text{enc}}), \\
\hat{h} &= W_{\text{dec}} z + b_{\text{dec}}.
\end{align}
We construct features that ``detect'' concepts, i.e., $s(h)=d^\top h + c$.

\paragraph{Lemma 3 (binding detector)} Let $u_j$ be the unbinding vector for role $r_j$ and let $f_\star$ be a filler embedding. Define
\begin{equation}
d_{(f_\star,r_j)} \;=\; (W^+)^\top (u_j \otimes f_\star), \qquad
c_{(f_\star,r_j)} \;=\; -d_{(f_\star,r_j)}^\top b.
\end{equation}
Then the scalar
\begin{equation}
s_{(f_\star,r_j)}(h) \;=\; d_{(f_\star,r_j)}^\top h + c_{(f_\star,r_j)}
\end{equation}
approximates the binding score $f_\star^\top E u_j$.

\emph{Proof.} Using the TPE-constructed linear probe weights (Equation~\ref{eq:tpe-probe}), and setting the set of filler embeddings $F$ as a single filler $f_\star$,
\begin{align}
f_\star^\top E u_j
&= (u_j \otimes f_\star)^\top W^+(h-b)
 \;=\; d_{(f_\star,r_j)}^\top h + c_{(f_\star,r_j)}.
\end{align}

\paragraph{Embedding into SAE parameters} For a feature with detector $(d,c)$, set the $k$-th encoder row to $W_{\text{enc}}[k,:]=d^\top$ and bias $b_{\text{enc}}[k]=c$. Then the SAE feature is $\text{ReLU}(d^\top h + c)$, which gives the binding score $f_i^\top E u_j$ for the filler $i$- role $j$ pair. This allows the SAE to reconstruct by setting $W_{dec} = W_{enc}^+$ and setting decoder bias such that it cancels the encoder bias, i.e. $b_{dec} = -W_{enc}b_{enc}$. The construction yields explicit SAE features for a collection of specific filler-role bindings derived from the TPE projection and unbinding vectors.

\section{The synthetic sequences task}

\subsection{Task and dataset}
\label{ap:seqs}

The sequence manipulation task is a sequence-to-sequence task that involves manipulating randomly generated sequences of tokens. We describe the complete data schema and generation procedure below.

We specify a vocabulary of 20 tokens. For clarity, we identify tokens with their index, $\{0, 1, 2 ..., 18, 19\}$. Each unit of the sequence is chosen independently from a uniform distribution over the vocabulary. Then, depending on the task, the target output is generated by algorithmically manipulating the input: for the copy task, it is the same as the input sequence; for reverse, the positions of tokens are reversed. We give the two tasks and an example input--output pair of each task below in Table \ref{tab:digit-eg}. A total of 50,000 examples, all of length 6, are generated for each task. They are split into 80\% (40,000) train, 10\% (5,000) evaluation, and 10\% (5,000) test. The generated sequences are cached and preserved across training runs.

\begin{table}[h]
    \centering
    \caption{Example input--output pairs for the sequence manipulation tasks: \emph{copy} returns the input unchanged, and \emph{reverse} outputs the reversed sequence.}
    \begin{tabular}{lcc}
        \toprule
        Task & Input & Output \\
        \midrule
        Copy    & \texttt{2 1 7 5 10 5} & \texttt{2  1 7 5 10 5} \\
        Reverse & \texttt{2 1 7 5 10 5} & \texttt{5 10 5 7 1  2} \\
        \bottomrule
    \end{tabular}
    \label{tab:digit-eg}
\end{table}

\subsection{Sequence-to-sequence models}
\label{ap:seqs-model-spec}

We train one-layer encoder--decoder RNNs, GRUs, and LSTMs on the two sequence tasks (copy, reverse) using the dataset described in Appendix~\ref{ap:seqs}. The trained model checkpoints for each task/architecture are then reused across all sequence experiments (TPE approximation, linear probes, and analogies). 

\paragraph{Special tokens} To format the task in a way that is friendly to encoder--decoder models, we prepend and append special tokens to the sequences. For the encoders, each input sequence is prepended with \texttt{<bos>} and appended with \texttt{<sep>}. For example, the input sequence \texttt{2 1 7 5 10 5} translates to 8 input tokens to the encoder, \texttt{<bos> 2 1 7 5 10 5 <sep>}. The decoder starts generation with \texttt{<bos>} as input, and we take the decoder's generation to be complete when it reaches 10 tokens or outputs an \texttt{<eos>} token.

\paragraph{Hyperparameters}

The models are all one-layer, with an embedding size of 64 and hidden state size of 256. The LSTM also has a cell state size of 256. The models do not have dropout. They are trained with the AdamW optimizer, with a batch size of 128, for 60 epochs. The learning rate follows a linear warmup over the first 100 steps, followed by a cosine decay schedule, peaking at 0.002. The best checkpoint, selected via validation split accuracy, is stored.

\paragraph{Evaluation and results}

The trained models are then evaluated for their accuracy on the test split. We compute two metrics: the \emph{token} level accuracy, which is teacher-forced by having the decoder receive the ground-truth input at each step, and computing the proportion of times the decoder outputs the correct token. The \emph{sequence} level accuracy uses no teacher forcing, instead letting the decoder generate until it emits an \texttt{<eos>} token or reaches 10 tokens, and computes the proportion of times the generated sequence matches the ground truth as a whole. The \emph{token} accuracy setting mirrors training more closely, while the \emph{sequence} accuracy measures the ability of the trained network to actually perform its training task. Both are reported in Table~\ref{tab:digits-task-accuracies-s}, and we see that the trained networks all achieve perfect accuracy.

\begin{table}[h]
  \centering
  \caption{Sequence-to-sequence model test accuracies with token- and sequence-level metrics.}
  \label{tab:digits-task-accuracies-s}
  \begin{tabular}{lcccccc}
    \toprule
    & \multicolumn{2}{c}{RNN} & \multicolumn{2}{c}{LSTM} & \multicolumn{2}{c}{GRU} \\
    \cmidrule(lr){2-3}\cmidrule(lr){4-5}\cmidrule(lr){6-7}
    Task & Token & Sequence & Token & Sequence & Token & Sequence \\
    \midrule
    Copy    & 1.0000 & 1.0000 & 1.0000 & 1.0000 & 1.0000 & 1.0000 \\
    Reverse & 1.0000 & 1.0000 & 1.0000 & 1.0000 & 1.0000 & 1.0000 \\
    \bottomrule
  \end{tabular}
\end{table}

\subsection{Tensor Product Encoders}
\label{ap:seqs-tpe-spec}

For each sequence-to-sequence model, we train a TPE to reconstruct the encoder's final hidden state. These TPEs are then stored and reused across the experiments: analogy evaluation (Appendix~\ref{ap:analogy}) and probe construction (Appendix~\ref{ap:probe-tpe-derivation}).

The output dimension of the TPEs trained to approximate RNNs and GRUs is 256. For LSTMs, the TPE approximates the \emph{concatenated} vector composed of the encoder's hidden state and cell state, so its output dimension is 512. The filler embeddings for the TPEs are the same size as the sequence-to-sequence model's vocabulary size. For roles, we do a simple left-to-right assignment of roles, including special tokens. After adding in a null token used for padding, this results in a total of 9 role embeddings. 

\paragraph{Hyperparameters} After doing an exploratory analysis optimizing for a large range of hyperparameters, we find that TPE training is quite insensitive to hyperparameters. We thus heuristically choose one set of hyperparameters proven to be successful approximating some of the sequence-to-sequence models, and reuse across all tasks and architectures. All the TPEs have filler and role dimensions 64, and are trained with batch size 64, learning rate 0.002 with a constant learning rate schedule, for 20 epochs. The checkpoint achieving the lowest evaluation loss during training is saved.

\paragraph{Evaluation and results} As introduced in Section~\ref{sec:tprs-tpe}, we evaluate trained TPEs via two metrics: the $R^2$ between the sequence-to-sequence model's hidden state and the TPE's approximation, and the substitution accuracy, which measures the accuracy of the decoder model if its initial hidden state is generated by the TPE instead of the encoder model. The results are reported in Table~\ref{tab:digits-task-accuracies-t}, where we see that all the TPE approximations achieve high $R^2$ and near-perfect substitution accuracies across both tasks.

\begin{table}[t]
  \centering
  \small
  \caption{Tensor Product Encoder substitution $R^2$ and token and sequence-level accuracies on the test split.}
  \label{tab:digits-task-accuracies-t}
  \begin{tabular}{lccccccccc}
    \toprule
    & \multicolumn{3}{c}{RNN} & \multicolumn{3}{c}{LSTM} & \multicolumn{3}{c}{GRU} \\
    \cmidrule(lr){2-4}\cmidrule(lr){5-7}\cmidrule(lr){8-10}
    Task & Token & Sequence & $R^2$ & Token & Sequence & $R^2$ & Token & Sequence & $R^2$ \\
    \midrule
    Copy    & 1.0000 & 1.0000 & 0.9853 & 1.0000 & 1.0000 & 0.9754 & 0.9998 & 0.9990 & 0.9343 \\
    Reverse & 1.0000 & 1.0000 & 0.8325 & 1.0000 & 0.9998 & 0.9721 & 0.9999 & 0.9996 & 0.9311 \\
    \bottomrule
  \end{tabular}
\end{table}

\begin{table}[t]
\centering
\caption{Probe accuracy (\%) comparison between trained vs analytically constructed probes on the sequence manipulation task. The results provided here correspond to the data used in plotting figure~\ref{fig:probes}.}
\setlength{\tabcolsep}{6pt}
\begin{tabular}{lcrrrr}
\toprule
\multirow{2}{*}{Model} & \multirow{2}{*}{Position} & \multicolumn{2}{c}{Copy} & \multicolumn{2}{c}{Reverse} \\
 & & Trained & Constructed & Trained & Constructed \\
\midrule
GRU & 1 & 100.00 & 100.00 & 58.70 & 57.34 \\
    & 2 & 100.00 & 100.00 & 85.84 & 76.68 \\
    & 3 & 99.88 & 99.96 & 91.52 & 84.70 \\
    & 4 & 99.12 & 99.08 & 98.40 & 97.32 \\
    & 5 & 94.32 & 86.84 & 100.00 & 99.96 \\
    & 6 & 95.00 & 90.18 & 100.00 & 100.00 \\
\midrule
LSTM & 1 & 100.00 & 100.00 & 65.86 & 75.62 \\
     & 2 & 97.78 & 95.28 & 70.58 & 63.48 \\
     & 3 & 91.12 & 86.70 & 75.36 & 68.24 \\
     & 4 & 86.74 & 88.12 & 87.12 & 82.68 \\
     & 5 & 85.14 & 71.68 & 98.90 & 98.82 \\
     & 6 & 96.20 & 98.62 & 100.00 & 100.00 \\
\midrule
RNN & 1 & 100.00 & 99.28 & 100.00 & 98.66 \\
    & 2 & 100.00 & 95.28 & 100.00 & 98.74 \\
    & 3 & 99.78 & 93.22 & 100.00 & 98.60 \\
    & 4 & 99.86 & 95.56 & 100.00 & 98.38 \\
    & 5 & 100.00 & 99.50 & 100.00 & 98.70 \\
    & 6 & 100.00 & 100.00 & 100.00 & 97.74 \\
\bottomrule
\end{tabular}
\label{tab:digits-probes}
\end{table}

\subsection{Linear probes}
\label{ap:seqs-probe-spec}

We train linear probes that take in the final encoder hidden state to predict the token identity at each input position. This results in 6 probes (one for each possible position) for each of the 6 sequence-to-sequence models.

The probes are all trained using the train split of the synthetic sequences dataset, with labels generated by algorithmically identifying the token that takes a specific position. 

\paragraph{Hyperparameters} All 6 (number of positions/probes per model) $\times$ 2 (number of tasks) $\times$ 3 (number of architectures) probes share the same hyperparameters: they are trained for 5 epochs, with a batch size of 256, constant learning rate of $5 \times 10^{-3}$.

\paragraph{Evaluation and results} We evaluate the probes on the test split and report accuracy as the proportion of times the probe's most likely label matches the ground truth. The accuracies for trained probes are presented as blue bars in Figure \ref{fig:probes}. The exact values can be found in Table~\ref{tab:digits-probes}.

\section{The SVO sentences task}

\subsection{Structured sentences}
\label{ap:sentences}

For transformer language models, we base our experiments on a set of simple subject-verb-object (SVO) sentences of the following structure:

\begin{center}
    \texttt{the <subject> will <verb> the <object>.}
\end{center}

To generate the dataset, we chose a list of 77 nouns of occupations (e.g. lawyer, student, farmer), and 5 verbs, ``see'', ``help'', ``visit'', ``teach'', ``call''. We selected these verbs because an occupation term is semantically reasonable as the subject or direct object for any of them. The dataset is generated by exhaustive combination of the above, creating a total of 77$^2$ $\times$ 5 $=$ 29645 examples. They are then divided into train, evaluation, and test splits at the ratio of 80\%:10\%:10\%, similar to the examples in the sequence task, resulting in 23,717 train examples, 2,965 evaluation examples, and 2.966 test examples. Tensor Product Encoders are trained with embeddings of sentences on the train split, selected on the evaluation split, and final metrics are reported on the test split. Since the dataset contains no duplicate sentences, all test sentences are unseen. Table \ref{tab:svo-examples} shows a few sample sentences generated by this procedure. The full list of occupations can be seen in Table \ref{tab:occupations}.

\begin{table}
    \caption{Example generated sentences.}
    \centering
    \renewcommand{\arraystretch}{1.05}
    \begin{tabular}{l}
        \hline
        \texttt{the carpenter will see the magician.} \\
        \texttt{the tutor will help the engineer.} \\
        \texttt{the violinist will see the dancer.} \\
        \texttt{the umpire will help the librarian.} \\
        \texttt{the intern will visit the scientist.} \\
        \hline
    \end{tabular}
    \label{tab:svo-examples}
\end{table}

\begin{table}
    \centering
    \setlength{\tabcolsep}{3.2pt} 
    \renewcommand{\arraystretch}{0.95} 
    \small
    \caption{List of the 77 occupation nouns used as \texttt{<subject>} and \texttt{<object>} in the SVO template.}
    \begin{tabular}{llllllll}
        \hline
        professor & student & president & judge & senator & secretary & doctor & lawyer \\
        scientist & banker & tourist & artist & author & actor & athlete & teacher \\
        engineer & accountant & architect & chef & photographer & farmer & ambassador & astronaut \\
        astronomer & blacksmith & baker & barber & biologist & butler & chemist & composer \\
        cartoonist & coach & captain & carpenter & dancer & director & drummer & detective \\
        explorer & economist & editor & governor & gardener & illustrator & intern & inventor \\
        journalist & linguist & manager & magician & mayor & miner & mathematician & musician \\
        novelist & nurse & painter & philosopher & physicist & politician & programmer & pilot \\
        poet & reporter & referee & sailor & spy & translator & treasurer & technician \\
        tutor & umpire & violinist & writer & librarian &  &  &  \\
        \hline
    \end{tabular}
    \label{tab:occupations}
\end{table}

\subsection{Tensor Product Encoders}
\label{ap:svo-tpe-spec}

We train TPEs to reconstruct sentence embeddings for three embedding models. The TPEs receive as input 3 filler-role pairs, corresponding to the subject, verb, and object of the sentence, respectively. As an example, the sentence \texttt{the carpenter will see the magician.} will be translated, algorithmically, into the TPE input pairs \{(\texttt{carpenter}, subject), (\texttt{see}, verb), (\texttt{magician}, object)\}. This filler-role scheme captures the possible variations between different sentences. The period and the words \textit{the} and \textit{will} are constant across all sentences in the dataset, so we hypothesize that they do not need to be included in the representation of each sentence because the bias term (of Equation~\ref{eq:tpe}) should be able capture the invariant offset that would result from including them. 

\paragraph{Encoding}

We use the SentenceTransformers library \cite{reimers-2019-sentence-bert} to access the embedding models and encode sentence representations. All models are loaded with \texttt{bfloat16} datatype. For ModernBERT, we use its embedding model variant \href{https://huggingface.co/nomic-ai/modernbert-embed-base}{\texttt{nomic-ai/modernbert-embed-base}}. The model is trained for vector embedding search, and requires an embedded sentence to be either a query or document. To align with its expected usage, we prepend ``\texttt{query\_document}'' to all sentences before encoding them. Qwen3-Embedding-8B and EmbeddingGemma encode sentences out-of-the-box, so we directly use them to encode the sentences with no modifications.

For the LLM experiments, we use decoder-only language models and cache the final-layer hidden state at the sentence-final period token via Huggingface Transformers. We use this punctuation-token representation because it occurs after the full SVO sentence has been processed.

\paragraph{Hyperparameters} We train TPEs to approximate the sentence embeddings previously generated with the hyperparameters listed below. 

\begin{enumerate}
    \item \textbf{ModernBERT:} Filler dimension 128, role dimension 4; trained for 100 epochs with an initial learning rate of 0.002, following cosine decay schedule.
    \item \textbf{Qwen3-Embedding-8B:} Filler dimension 128, role dimension 4; trained for 100 epochs with an initial learning rate of 0.002, following cosine decay schedule.
    \item \textbf{EmbeddingGemma:} Filler dimension 128, role dimension 4; trained for 100 epochs with an initial learning rate of 0.005, following cosine decay schedule.
\end{enumerate}

For the LLM punctuation-token TPEs, we use the following hyperparameters:

\begin{enumerate}
    \item \textbf{Qwen3-8B:} Filler dimension 256, role dimension 4; trained for 100 epochs with an initial learning rate of 0.004, following a cosine decay schedule.
    \item \textbf{OLMo-2-13B:} Filler dimension 256, role dimension 4; trained for 100 epochs with an initial learning rate of 0.004, following a cosine decay schedule.
    \item \textbf{GPT-OSS-20B:} Filler dimension 256, role dimension 4; trained for 100 epochs with an initial learning rate of 0.004, following a cosine decay schedule.
\end{enumerate}

We evaluate TPEs at the end of every epoch on the evaluation split of the dataset, and save when the metrics are better than the last saved checkpoint. For each model, we choose the checkpoint with best evaluation loss and report its performance on the test split. The final metrics for all TPEs are reported in Table~\ref{tab:svo-tpe-reconstruction}

\begin{table}
\centering
\caption{SVO TPE metrics on the held-out test split.}
\begin{tabular}{llrr}
\toprule
Type & Model & $R^2$ & MSE \\
\midrule
\multirow{3}{*}{Embedding Model} & ModernBERT & 0.9396 & $4.27 \times 10^{-5}$ \\
 & EmbeddingGemma & 0.9155 & $1.57 \times 10^{-5}$ \\
 & Qwen3-Embedding-8B & 0.9000 & $9.51 \times 10^{-6}$ \\
\midrule
\multirow{3}{*}{LLM} & Qwen3-8B & 0.7052 & $3.50 \times 10^{-2}$ \\
 & OLMo-2-13B & 0.6440 & $3.28 \times 10^{-2}$ \\
 & GPT-OSS-20B & 0.7423 & $5.16 \times 10^{-1}$ \\
\bottomrule
\end{tabular}
\label{tab:svo-tpe-reconstruction}
\end{table}

\begin{table}
\centering
\caption{SVO probe accuracy (\%) for trained vs analytically constructed probes on SVO sentences.}
\begin{tabular}{llrr}
\toprule
Model & Role & Trained & Constructed\\
\midrule
ModernBERT & Subject & 99.26 & 89.61 \\
          & Verb & 100.00 & 98.01 \\
          & Object  & 96.49 & 91.53 \\
\midrule
EmbedGemma & Subject & 99.83 & 100.00 \\
          & Verb & 100.00 & 99.83 \\
          & Object  & 99.87 & 99.80 \\
\midrule
Qwen3-8B   & Subject & 99.93 & 99.70 \\
          & Verb & 100.00 & 100.00 \\
          & Object  & 99.87 & 98.31 \\
\bottomrule
\end{tabular}
\label{tab:sentences-probe-svo}
\end{table}

\subsection{Linear probes}
\label{ap:svo-probe-spec}

We train linear probes to classify the subject, verb, and object given a sentence embedding. Probes are trained with all 23,717 sentences in the train set, batch size 256, and a constant learning rate of 0.005. We train for 20 epochs for ModernBERT and EmbeddingGemma, and 50 epochs for Qwen3-Embedding-8B, confirming that evaluation loss plateaus by inspecting training logs. Due to the simplicity of the task, we did not perform automated hyperparameter searches. Analytic probes are evaluated with batch size 256 on the held-out test split. Results are reported for each probe (subject/verb/object) on the test split.

Each role probe predicts a 77-way subject/object label or a 5-way verb label. All trained probes perform well, and all but the object probe trained on ModernBERT reach near-perfect accuracy. The results are visualized in Figure~\ref{fig:probes}b, and also presented in Table~\ref{tab:sentences-probe-svo}.

\section{Additive analogy construction and evaluation}
\label{ap:analogy}

Below we describe the exact procedure for constructing and evaluating additive analogy quartets on our task setups.

\subsection{Synthetic sequences}
\label{ap:analogy-seqs}

The analogy quartets are set up as follows. Denote the 4 sequences of an analogy quartet $A, B, C, D$. We draw sequence $A$ from the test split of each task. Given $A=(a_1,\dots,a_6)$ (since all sequences in the dataset are length 6), we sample a set of \emph{substituted roles} $S \subset \{1,\dots,6\}$ uniformly at random with $1 \le |S| \le 3$. We call the remaining positions, $R=\{1,\dots,6\}\setminus S$, the \emph{retained roles}. We then construct sequences
\[
B = A^{(R \leftarrow \tilde{a})},
C = A^{(S \leftarrow \hat{a},\: R \leftarrow \tilde{a})},
D = A^{(S \leftarrow \hat{a})},
\]
where $S^{(P \leftarrow a)}$ denotes the \emph{replacement} operation, replacing the positions in $P$ with random tokens from $\hat{a}$ and $\tilde{a}$, two randomly generated sequences that are different at each position from the original sequence $A$. Thus, $B$ and $C$ differ only on $S$, and $D$ is $A$ with exactly those positions replaced. The analogy quartets all contain special tokens, so the recurrent encoders can encode them as normal inputs. Example~\ref{ex:seqs-analogy} demonstrates a quartet constructed from the algorithm above:

\ex. \textbf{Example (sequence analogy):}\label{ex:seqs-analogy}\\
$A=$ \text{\texttt{<bos> 2 1 7 5 10 5 <sep>}} \\
$B=$ \text{\texttt{<bos> 9 3 7 8 10 4 <sep>}} \\
$C=$ \text{\texttt{<bos> 9 3 0 8 10 4 <sep>}} \\
$D=$ \text{\texttt{<bos> 2 1 0 5 10 5 <sep>}}

We evaluate analogies by sampling 1000 analogies per model and task. We compute analogy vectors using both the standard three-embedding method and the TPE-constructed method, then evaluate top-$k$ accuracy using the algorithm described in Appendix~\ref{ap:analogy-eval}.

\subsection{SVO sentences}
\label{ap:analogy-svo}

For SVO sentences, we construct analogies from the test split by varying exactly one role (subject, verb, or object) while holding the other two fixed. For \emph{subject} analogies, we group sentences by fixed (verb, object) values; within each group, for every ordered pair of distinct subjects $(\mathrm{val}_A,\mathrm{val}_C)$, we seek another group with a different (verb, object) pair that contains both subjects. We then form $A$ and $D$ from the first group and $B$ and $C$ from the second group, yielding a quartet $A, B, C, D$ where only the subject role varies. Object analogies are constructed analogously by grouping on (subject, verb) and varying the object role. We stop at the first matching second group for each $(\mathrm{val}_A,\mathrm{val}_C)$ pair.

Analogies are ranked against all possible SVO sentences, although the quadruples themselves are built only from the test split. This evaluation procedure yields exactly 44,468 analogies in total (22,488 subject analogies and 21,980 object analogies). Below we show one such example quartet:

\ex. \textbf{Example (subject analogy):}\\
$A=$ \text{``the doctor will see the patient .''} \\
$B=$ \text{``the doctor will help the student .''} \\
$C=$ \text{``the lawyer will help the student .''} \\
$D=$ \text{``the lawyer will see the patient .''}

\subsection{Evaluation}
\label{ap:analogy-eval}

\paragraph{Top-$k$ accuracy} Given an analogy vector $v$ (either $e(A)-e(B)+e(C)$ or the TPE-constructed variant) and a candidate pool of embeddings $\{e_i\}_{i=1}^N$, we rank candidates by cosine similarity
\[
s_i \;=\; \frac{v^\top e_i}{\|v\|_2\,\|e_i\|_2}.
\]
Let $i^\star$ be the index of the target sentence/sequence $D$. The rank of the target is
\[
\mathrm{rank}(v) \;=\; 1 + \sum_{i=1}^N \mathbbm{1}\{s_i > s_{i^\star}\},
\]
and top-$k$ accuracy is the fraction of analogies with $\mathrm{rank}(v) \le k$.

\paragraph{A parallelized implementation} Computing cosine similarity between vectors and ranking them sequentially is inefficient and takes an unreasonable amount of time even for our relatively small evaluation set. We implement the same computation in a vectorized way. First, cosine similarities are computed in batch between all embeddings. Then, we index the target's similarity as an entry from the resulting cosine similarity matrix, and do distributed comparison between the target cosine similarity and every cosine similarity with masking. We present implementation pseudocode below:

\begin{algorithm}
\caption{Computing top-$k$ accuracy for analogies}
\begin{lstlisting}
# inputs:
#   V: analogy vectors, shape [B, D]
#   E: candidate embeddings, shape [N, D]
#   target_idx: index of target D for each analogy, shape [B]
#   k: specifies top-k accuracy
# obtains:
#   topk: the top-k accuracy

E = E / (norm(E, axis=1, keepdims=True) + 1e-12)   # [N, D]
V = V / (norm(V, axis=1, keepdims=True) + 1e-12)   # [B, D]

S = E @ V.T                                         # [N, B] Cosine similarity matrix
target_sim = S[target_idx, arange(B)]               # [B]
ranks = 1 + (S > target_sim).sum(axis=0)            # [B]

topk = mean(ranks <= k)
\end{lstlisting}
\end{algorithm}

\section{Training and evaluating sparse autoencoders}
\label{ap:sae-training}

We train SAEs on SVO sentence representations using two variants: top-$k$ SAEs, and supervised SAEs, where the first is a standard performant SAE architecture, and the second is a modified architecture specifically suiting our purposes. For embedding models, the representations are sentence embeddings. For language models, the representations are cached final-layer hidden states at the sentence-final period token, using the same punctuation-token representations described in Appendix~\ref{ap:svo-tpe-spec}. All SAEs are trained on the train split of the relevant SVO dataset, then evaluated on the held-out test split.

Throughout this section, we term a row in the SAE's weight encoder weight matrix a ``feature'', and the value corresponding to it in the intermediate layer of the SAE the ``feature activation''. We term the explanation assigned to the feature the ``concept'', which is often a semantic content-syntactic role pair, such as ``subject-linguist''.

\subsection{Reconstruction $R^2$} 

We report reconstruction $R^2$ and feature quality. For reconstruction, given inputs $x$ and reconstructions $\hat{x}$, we compute
\begin{equation}
R^2 = 1 - \frac{\sum \|x-\hat{x}\|_2^2}{\sum \|x-\bar{x}\|_2^2},
\end{equation}
where $\bar{x}$ is the mean representation in the dataset used for evaluation. $R^2$ captures how well the SAE's reconstruction recovers the input representations with respect to the total variations within the evaluation dataset, where 0 stands for (roughly speaking) no correlation, and 1 stands for perfect reconstruction. This metric allows us to quantify the quality of SAE reconstructions across models that widely differ in sizes. 

\subsection{Feature quality}
\label{ap:sae-feature-quality}

\paragraph{Motivation} A feature is meaningful if its activations correlate well with whether a given concept is present within the example --- intuitively, the feature should activate highly when the concept it tracks is present, and should not activate (as highly) when the concept is not. We cash out this notion of ``well-rankedness'' by  asking, for a given feature and concept, the probability that a positive example (containing the concept) is ranked above a negative example (not containing the concept) by that feature’s activation. 

\paragraph{Definition} For each feature $k$, let $a_i^{(k)}$ denote its activation on example $i$. Fix a concept $y_k$ for this feature and define the positive and negative sets
\[
P_k=\{i:\text{example } i \text{ contains } y_k\},\qquad
N_k=\{i:\text{example } i \text{ does not contain } y_k\}.
\]
We define the quality score for the feature evaluated on this concept as the probability a randomly sampled pair of positive-negative samples are \emph{well-ranked}, i.e.,
\[
\mathrm{Quality}(k)=\frac{1}{|P_k| |N_k|}\!\sum_{i\in P_k}\sum_{j\in N_k}
\Big[\mathbbm{1}\!\{a_i^{(k)} > a_j^{(k)}\} + \tfrac{1}{2}\mathbbm{1}\!\{a_i^{(k)} = a_j^{(k)}\}\Big].
\]
This metric gives $1.0$ for perfect separation, and $0.0$ for perfectly reversed ranking. To compute the quality score for an SAE, we take the average across the quality score of all its features. The pseudocode illustrating how we compute feature quality of an SAE in the code implementation is presented below:

\begin{algorithm}
\caption{Computing feature quality}
\begin{lstlisting}
# inputs:
#   A: activations, shape [num_examples, num_features]
#   example_labels: concepts that are present in each example (e.g., subject-linguist)
#   feature_labels: assigned concept for each feature
# obtains:
#   scores: list of quality scores for every feature

scores = []
for k in range(num_features):
    y = feature_labels[k]
    pos = where(example_labels == y)
    neg = where(example_labels != y)
    if len(pos) == 0 or len(neg) == 0:
        continue
    a_pos = A[pos, k]   # [n_pos]
    a_neg = A[neg, k]   # [n_neg]
    score = mean( (a_pos[:,None] > a_neg[None,:]) +
                  0.5 * (a_pos[:,None] == a_neg[None,:]) )
    scores.append(score)
\end{lstlisting}
\end{algorithm}

\paragraph{Assigning concept to unsupervised SAEs} For supervised and constructed SAEs, each feature is explicitly trained to correspond to a pre-defined concept. However, top-$k$ SAEs are unsupervised, and we do not know a priori what concept a feature corresponds to, if anything. We thus run the above evaluation process for each feature with every possible concept on the train set, and assigning the feature the concept for which it performs best on. We then report the quality score by evaluating it again on the test set, this time specifying the concept as the one we previously assigned.

\subsection{Top-$k$ SAEs}
\label{ap:sae-training:topk}

We train top-k SAEs \cite{makhzani2014ksparseautoencoders}, which can be thought of as a way to modify standard SAEs by replacing the activation function with keeping only the $k$-highest activated units.

As there is no guarantee that the $k$-highest activated units follow the same distribution between training and test stages, we follow the standard procedure \cite{gao2024scalingevaluatingsparseautoencoders} of storing a moving average of the numerical threshold for top-$k$ activations during training, and use the numerical threshold instead at evaluation time.

Following standard practices \cite{templeton2024scaling}, we also employ \emph{dead neuron resampling}: with randomly initialized features, it may be the case that some feature directions never get activated by examples in the training set. We thus keep a running record of features that have not been activated during training (or since the last resampling step), and resample them by randomly initializing them again. During training, we set an \emph{n resamples} hyperparameter, which then defines evenly spaced resampling steps throughout training. For example, training for 20 epochs with 4 resampling steps will result in resampling at the end of the \nth{4}, \nth{8}, \nth{12}, and \nth{16} epochs. We have found that increasing the number of times resampling is performed to be essential for the SAEs to discover high-quality features that align with human expectations (e.g., subject-doctor).

\paragraph{Hyperparameters} Due to standard concerns about hyperparameter sensitivity during SAE training \cite{paulo2025sparseautoencoderstraineddata}, we perform a random search over the span of hyperparameters: $k \in \{4, 8, 16, 32\}$, hidden dimension $\in \{128, 256, 512, 1024\}$, batch size $\in \{64, 128, 256\}$, learning rate $\in [10^{-5}, 5\times 10^{-3}]$ (sampled with log-uniform distribution), and resample times $\in \{1,2,5\}$.

We train each model for 200 epochs,\footnote{This is a significantly larger number of epochs compared to what is commonly seen for SAE training, and we find this to improve results significantly as our training dataset is quite small (approximately 25 thousand examples).} where the first 5\% (10 epochs) applies linear learning rate warmup, and heuristically select the checkpoint (among all hyperparameter search  checkpoints) that achieves the best sparsity-reconstruction tradeoff: that is, the trained SAE should have sparsely activated features with a sparsity that aligns with human expectations. Any SAE that is more sparse will have a worse loss than the chosen one, and any SAE with lower loss will not be as sparse. In the end, the chosen top-$k$ SAE checkpoints have the following hyperparameters:

\begin{itemize}
    \item ModernBERT: hidden dimension $1024$, $k = 8$. Learning rate $= 6.658 \times 10^{-4}$, batch size $= 256$.
    \item Qwen3-Embedding-8B: hidden dimension $1024$, $k = 32$. Learning rate $= 1.508 \times 10^{-3}$, batch size $= 256$.
    \item EmbeddingGemma: hidden dimension $1024$, $k = 8$. Learning rate $= 4.815 \times 10^{-4}$, batch size $= 128$.
\end{itemize}

For the language-model punctuation-token experiments, the top-$k$ baselines are selected from 40 total sweeps run over hidden dimension $\{512, 1024, 2048, 4096\}$, $k \in \{8, 16, 32, 64, 128\}$, batch size $\{128, 256\}$, learning rate sampled log-uniformly from $[10^{-5}, 5\times 10^{-3}]$, and resampling times $\{2, 5, 8\}$. The sweep uses Weights and Biases' Bayesian backend. These models are also trained for 200 epochs. The selected checkpoints optimize feature quality on the validation split, with the following hyperparameters:

\begin{itemize}
    \item Qwen3-8B: hidden dimension $4096$, $k = 64$. Learning rate $= 1.69 \times 10^{-5}$, batch size $= 256$, resample times $=5$.
    \item OLMo-2-13B: hidden dimension $2048$, $k = 128$. Learning rate $= 3.72 \times 10^{-3}$, batch size $= 128$, resample times $=2$.
    \item GPT-OSS-20B: hidden dimension $1024$, $k = 128$. Learning rate $= 4.19 \times 10^{-3}$, batch size $= 128$, resample times $=5$.
\end{itemize}

\paragraph{Supervised SAEs} A supervised SAE pre-assigns a concept for each of its features at initialization time, and as such always has 159 features (77 (subjects) + 77 (objects) + 5 (verbs)). In addition to standard SAE losses, it adds an auxiliary ``supervision loss'', which is a cross-entropy loss between the SAE's pre-ReLU intermediate activations and the ground truth concepts that are present. Each sentence in our dataset contains 3 concepts (one each for subject, verb, and object), so the target distribution the intermediate activations are trained to approximate is one that assigns three labels $1/3$ probability, and 0 to the rest of the labels. Formally, given an input embedding $x \in \mathbb{R}^d$, the encoder computes pre-activations and activations
\begin{equation}
a = W_e x + b_e, \qquad z = \mathrm{ReLU}(a),
\end{equation}
and the decoder reconstructs
\begin{equation}
\hat{x} = W_d z + b_d .
\end{equation}
The standard SAE loss is the sum of reconstruction error and an $L_1$ sparsity
penalty on the activations:
\begin{equation}
\mathcal{L}_{\mathrm{SAE}} = \| \hat{x} - x \|_2^2 + \lambda \| z \|_1 ,
\end{equation}
where $\lambda$ is the sparsity weight. The supervision loss is applied to the
encoder \emph{pre-activations} $a$. For each example we build a multi-label
target vector $y \in \{0,1\}^F$ with three ones (subject, verb, object) and
normalize it to a distribution $\tilde{y} = y / \sum_j y_j = y/3$. We then
compute a cross-entropy loss against the softmax of $a$:
\begin{equation}
\mathcal{L}_{\mathrm{sup}} = -\sum_{j=1}^F \tilde{y}_j \log \mathrm{softmax}(a)_j .
\end{equation}
The total objective is
\begin{equation}
\mathcal{L} = \mathcal{L}_{\mathrm{SAE}} + \alpha \mathcal{L}_{\mathrm{sup}},
\end{equation}
where $\alpha$ is the supervision weight.

As cross-entropy loss biases features towards negative infinity, we find that setting the supervision weight and no additional L1 regularization is sufficient for sparse activations. The embedding-model SAEs are trained with sparsity penalty 0.0, and supervision weight 0.05 (ModernBERT, EmbeddingGemma) or 0.02 (Qwen3-Embedding-8B). The language-model SAEs use sparsity penalty 0.0, supervision weight 0.02, and a dead-latent threshold of $10^{-6}$. All supervised SAEs are trained for 40 epochs, learning rate $2\times 10^{-3}$, warmup ratio 0.05, and batch size 128. For the language-model SAEs, we checked seeds $\{13,42,101\}$ and found stable reconstruction at the displayed precision: mean $R^2$ is 0.9979 for Qwen3-8B, 0.9894 for OLMo-2-13B, and 0.9907 for GPT-OSS-20B.

\section{TPR substitutions systematically generalize}
\label{ap:holdout}

TPRs have a compositional structure that is systematically constructed from filler and role embeddings. If this predicted structure matches with that of the neural network representations that we are analyzing, then we would expect TPR approximations to \emph{generalize}: that is, even if we train the TPR approximation on only a subset of possible filler-role combinations, the learned TPR approximation should nevertheless generalize to novel filler-role combinations in the evaluation set. Below we test this hypothesis on both the synthetic sequence tasks and the SVO sentences task, showing that in both situations the TPE approximation generalizes to novel filler-role pairs. 

\subsection{Synthetic Sequences}

For the synthetic sequences experiments, we construct fixed-length-6 copy and reverse datasets in which one randomly chosen number-position pair is held out at each position: $(2,1)$, $(7,2)$, $(13,3)$, $(4,4)$, $(18,5)$, and $(10,6)$. The train, validation, and test splits contain no examples with any of these held-out pairs, while the generalization split contains 5000 examples, each of which contains at least one held-out pair. Thus the generalization split tests whether a TPE trained without a particular filler-role binding can still reconstruct the encoder state for sequences containing that binding.

We reuse the same frozen sequence-to-sequence models as in Appendix~\ref{ap:seqs}. The TPEs are trained with the same hyperparameters as Appendix~\ref{ap:seqs-tpe-spec}. As before, we evaluate the TPEs with reconstruction $R^2$ and substitution accuracy, where substitution accuracy measures decoder accuracy when the encoder hidden state is replaced by the TPE approximation.

\begin{table}[h]
  \label{tab:digits-gen}
  \centering
  \small
  \caption{Digits filler-role holdout results. Test examples contain no held-out filler-role pairs; generalization examples contain at least one held-out filler-role pair.}
  \label{tab:digits-holdout-tpe}
  \begin{tabular}{llcccc}
    \toprule
    & & \multicolumn{2}{c}{Test} & \multicolumn{2}{c}{Generalization} \\
    \cmidrule(lr){3-4}\cmidrule(lr){5-6}
    Task & Model & Substitution Acc. & $R^2$ & Substitution Acc. & $R^2$ \\
    \midrule
    Copy & RNN  & 1.0000 & 0.9863 & 1.0000 & 0.9835 \\
     & LSTM & 1.0000 & 0.9949 & 0.9634 & 0.9882 \\
     & GRU  & 0.9994 & 0.9600 & 0.8230 & 0.9005 \\
    Reverse & RNN  & 1.0000 & 0.8344 & 1.0000 & 0.8238 \\
     & LSTM & 0.9994 & 0.9888 & 0.9464 & 0.9758 \\
     & GRU  & 0.9998 & 0.9497 & 0.9098 & 0.9000 \\
    \bottomrule
  \end{tabular}
\end{table}

As seen in Table~\ref{tab:digits-gen}, the held-out substitution results remain strong. Averaged across the six TPEs, the generalization split has $R^2=0.9287$ and substitution accuracy $0.9404$. The RNN TPEs show perfect accuracy on both the test and generalization splits. The GRU and LSTM TPEs show some performance drops, but still performs well ($\geq$90\%) on the OOD generalization set.

\subsection{SVO sentence embeddings}

We construct an analogous holdout split for the SVO sentence embeddings. Because verbs occur only in the verb role in this dataset, we hold out noun-role pairs rather than verb-role pairs. Seven nouns are held out in subject position (\texttt{professor}, \texttt{artist}, \texttt{ambassador}, \texttt{coach}, \texttt{gardener}, \texttt{musician}, and \texttt{referee}) and seven different nouns are held out in object position (\texttt{secretary}, \texttt{engineer}, \texttt{barber}, \texttt{drummer}, \texttt{linguist}, \texttt{physicist}, and \texttt{technician}). The train, validation, and test splits contain no sentence with these noun-role pairs, while the generalization split contains all sentences with at least one held-out noun-role pair. This yields 19,600 training sentences, 2450 validation sentences, 2450 test sentences, and 5145 generalization sentences.

We train TPEs with the same filler-role encoding, embedding models, and hyperparameters as Appendix~\ref{ap:svo-tpe-spec}. The evaluation again compares the TPE reconstruction to the frozen sentence embedding, but for this setting we report only reconstruction $R^2$.

\begin{table}
  \label{tab:svo-gen}
  \centering
  \small
  \caption{SVO sentence-embedding filler-role holdout reconstruction results.}
  \label{tab:svo-holdout-tpe}
  \begin{tabular}{lcc}
    \toprule
    Model & Test $R^2$ & Generalization $R^2$ \\
    \midrule
    ModernBERT & 0.9679 & 0.9589 \\
    EmbeddingGemma & 0.9883 & 0.9602 \\
    Qwen3-Embedding-8B & 0.9619 & 0.9350 \\
    \bottomrule
  \end{tabular}
\end{table}

As in the digits setting, TPE reconstruction generalizes to unseen filler-role combinations (Table~\ref{tab:svo-gen}). The average $R^2$ on the SVO generalization split is $0.9514$, compared to $0.9727$ on the in-distribution test split. This result indicates that the TPEs capture filler-role structures within the neural network activations that transfer to systematic substitutions. 



\end{document}